\documentclass[10pt,twocolumn,letterpaper]{article}

\usepackage{cvpr}
\usepackage{times}
\usepackage{epsfig}
\usepackage{graphicx}
\usepackage{amsmath}
\usepackage{amssymb}
\newcommand{\ignore}[1]{}

\newcommand{\ba}{\begin{array}}
\newcommand{\ea}{\end{array}}
\newcommand{\bc}{\begin{center}}
\newcommand{\ec}{\end{center}}
\newcommand{\be}{\begin{enumerate}}
\newcommand{\ee}{\end{enumerate}}
\newcommand{\bea}{\begin{eqnarray}}
\newcommand{\eea}{\end{eqnarray}}
\newcommand{\beas}{\begin{eqnarray*}}
\newcommand{\eeas}{\end{eqnarray*}}
\newcommand{\beq}{\begin{equation}}
\newcommand{\eeq}{\end{equation}}
\newcommand{\bfig}{\begin{figure}}
\newcommand{\efig}{\end{figure}}
\newcommand{\bi}{\begin{itemize}}
\newcommand{\ei}{\end{itemize}}
\newcommand{\bpic}{\begin{picture}}
\newcommand{\epic}{\end{picture}}
\newcommand{\btabular}{\begin{tabular}}
\newcommand{\etabular}{\end{tabular}}
\newcommand{\btable}{\begin{table}}
\newcommand{\etable}{\end{table}}

\newcommand{\es}{\vfill
                 \rule[-6mm]{170mm}{0.7mm} \\
                 \redw{{\tiny
		  \hfill S-\theslide}}
                 \end{slide}}

\newcommand{\matxx}[1]{{\mathtt #1}}
\newcommand{\vecXX}[1]{{\mathbf {#1}}}
\newcommand{\vecYY}[1]{{\boldsymbol {#1}}}

\providecommand{\etal}{{\em et~al.}}

\def \hbar {{\bar{h}}}

\def \vech {{\vecXX{h}}}

\def \vecx {{\vecXX{x}}}

\def \vecz {{\vecXX{z}}}

\def \vecX {{\vecXX{X}}}

\def \veceta   {{\vecYY{\eta}}}
\def \vecmu    {{\vecYY{\mu}}}

\def \matJ {{\matxx{J}}}

\def \matLambda {{\matxx{\Lambda}}}

\newcommand{\vecthree}[3]{\left(\begin{array}{c}#1\\#2\\#3\end{array}\right)}

\newcommand{\vecone}[1]{\left(\begin{array}{c}#1\end{array}\right)}

\newcommand{\vectwo}[2]{\left(\begin{array}{c}#1\\#2\end{array}\right)}

\newcommand{\rowtwo}[2]{\left(\begin{array}{cc}#1&#2\end{array}\right)}

\newcommand{\matthree}[9]{\left[\begin{array}{ccc}#1&#2&#3\\#4&#5&#6\\#7&#8&#9\end{array}\right]}

\newcommand{\mattwofour}[8]{\left[\begin{array}{cccc}#1&#2&#3&#4\\#5&#6&#7&#8\end{array}\right]}

\newcommand{\mattwo}[4]{\left[\begin{array}{cc}#1&#2\\#3&#4\end{array}\right]}

\newcommand{\matone}[1]{\left[\begin{array}{c}#1\end{array}\right]}

\makeatletter
\renewcommand*\env@matrix[1][*\c@MaxMatrixCols c]{%
  \hskip -\arraycolsep
  \let\@ifnextchar\new@ifnextchar
  \array{#1}}
\makeatother

\usepackage[pagebackref=true,breaklinks=true,letterpaper=true,colorlinks,bookmarks=false]{hyperref}

\cvprfinalcopy % Uncomment this line for the final submission

\ifcvprfinal\fi
\begin{document}

%%%%%%%%% TITLE
%\title{Smart pyramids for robust dense image alignment}
\title{FutureMapping 2: Gaussian Belief Propagation for Spatial AI}

\author{Andrew J. Davison and Joseph Ortiz\\
{\tt\small [a.davison,j.ortiz18]@imperial.ac.uk}\\
Department of Computing, Imperial College London, UK
}

%\tableofcontents

\maketitle
\thispagestyle{empty}

\begin{abstract}
We argue the case for Gaussian Belief
Propagation (GBP) as a strong algorithmic framework for the
distributed, generic and incremental probabilistic estimation we need
in Spatial AI as we aim at high performance smart robots and devices
which operate within the constraints of real products.
Processor hardware is changing rapidly, and GBP has the right
character to take advantage of highly distributed processing and
storage while estimating global quantities, as well as great
flexibility. We present a detailed tutorial on GBP, relating it to
the standard factor graph formulation used in robotics and computer
vision, and give several simulation examples with code which
demonstrate its properties.
  \end{abstract}

\section{Introduction}

Spatial AI is the real-time
vision-driven capability that robots and other devices need to
understand and interact intelligently with the spaces around them,
while satisfying the constraints such as power usage, compactness, robustness and simplicity enforced by real
products. Davison~\cite{Davison:ARXIV2018}, set out the case 
that there are still orders of magnitude of improvement needed
in efficient performance to deliver the capabilities needed for
breakthrough products such as lightweight home robots or AR glasses. 
Prototype real-time scene understanding systems in academia such as
SemanticFusion~\cite{McCormac:etal:ICRA2017} require heavy computing
resources while delivering a fraction of the capability needed.
While there is much ongoing effort in industry to optimise and
engineer such methods for embedded implementation, we believe that
there are many fundamental changes needed still
to cross this gap across algorithms, processors and sensors.

In this follow-on paper we present the case for Gaussian Belief
Propagation (GBP) as a very strong algorithmic framework for the
distributed, generic and incremental probabilistic estimation we need
in Spatial AI.

GBP is a special case of general Loopy Belief Propagation, where an
estimation problem represented by a factor graph can be solved
in an iterative manner by computation at the nodes of the
graph and purely local message passing between then.
It is at
heart a simple algorithm but can in our experience be subtle to
understand. Alongside broader discussion, we therefore give a very
detailed derivation of BP and GBP from first principles, and link it
directly with the factor graph and non-linear optimisation methods and
terminology commonly used in robotics and computer vision. We present
some demo implementations for 1D and 2D SLAM-like problems, with open
source Python code available for readers to experiment further.

GBP is not novel, and has even previously been tested in SLAM
settings~\cite{Ranganathan:etal:IJCAI2007}, but has not yet been
seriously used in practical Spatial AI problems. We believe that
recent advances in computing hardware in particular make this the
right time to re-evaluate its properties.

\subsection{Spatial AI and Computer Architecture}

Spatial AI at its core is a problem of incremental estimation, where a
persistent scene model, with static and dynamic elements, must be stored and updated continually using
data from various sources. Some data will be a flow of geometric measurements
from a metric sensor; other data  could be labelling output from a neural
network; and yet more could be prior information from assumptions made at the
start of the mapping process or communicated later on, such as  the
calibration parameters of a robot's drive system.
All of this data must be combined consistently into the chosen scene
representation, which could be complicated and heterogeneous, consisting
of multiple geometric and semantic representations such as meshes,
volumes, learned shape spaces, semantic label fields, or the estimated
locations of parametric CAD objects.

Current prototype Spatial AI systems, attempting to process this
heterogeneous flow of data into complicated persistent representations
via various estimation techniques, often have severe performance
bottlenecks due to limits in the capacities of computation load, data
storage or data transfer. We think that there are two promising and parallel lines of
attack to enable progress here. One is to focus on scene representation,
and to find new parameterisations of world models which allow high
quality scene models to be built and maintained much more
efficiently. The CodeSLAM~\cite{Bloesch:etal:CVPR2018} and SceneCode~\cite{Zhi:etal:CVPR2019} projects for instance are
steps in this direction, using deep learning to find coded, compressed
representations of geometry and semantic labels which can then be
optimised to fuse multi-view data.

The other is to look towards the changing landscape in computing and
sensing hardware.
(We highly recommend the recent PhD thesis of Julien Martel for ambitious thinking about this whole area \cite{Martel:PHD2019}).
Computing hardware is at the beginning of a revolution, as we move
away from reliance on processors and memory systems designed either for
completely general purpose use (CPUs) or computer graphics (GPUs)
towards an era where AI, and perhaps Spatial AI in particular, are
significant enough applications to drive the development of custom
computing hardware.

Fitting computer architecture to applications, in particular with the
aim of reducing power usage while maintaining performance, certainly
entails massive parallelism~\cite{Sutter:Jungle2011}; but also,
fundamentally, intermingling data storage and computation to reduce
the `bits $\times$ millimetres' through which data is moved.  GPUs are
now one of the main workhorses of AI computation, and are certainly
very good for certain tasks in computer vision, but we believe that
the future of Spatial AI compute will require much more flexible
storage and computation. This is particularly true due to the
closed-loop, incremental nature of Spatial AI, where new data must be
continually compared to and combined with stored models.
In SemanticFusion, a GPU is used for tasks such as CNN label
prediction and dense image alignment. In industry, many specialised
chips and architectures are emerging to accelerate important computer
vision algorithms such as feature detection and tracking or even a
whole visual odometry pipeline, and certainly there is a great deal of
effort on developing specific architectures for CNNs.
However, as examined in~\cite{Davison:ARXIV2018}, a complete Spatial
AI system requires 
many other computations, and current prototypes 
rely heavily on CPU work, main RAM
storage, and high bandwidth data transfers.

New graph processors designs such
as Graphcore's IPU~\cite{Graphcore} are emerging which have taken quite general design
choices towards enabling a different type of processing. The IPU is a
massively parallel chip, but where the processing cores are embedded
in a large amount of high performance on-chip memory which can be used
generally for local storage and inter-core communication.
In the IPU, computation works best when the storage it needs can be distributed 
around the chip close to the cores and there is no need ever to
communicate with external off-chip memory. However, the total on-chip
memory is relatively small compared to off-chip RAM so algorithmic
choices to `recompute instead of store' are advantageous.

Thinking even more generally, we can predict a future where many
intelligent devices operate in a coordinated way within a space, some
of them quite simple, and where efficiency in each device emphasizes
local computation (`edge compute') and minimal inter-device
communication. If these devices are to coordinate to estimate global
quantities, the computation must also be graph-based and distributed,
with local computation and storage.

\subsection{Probabilistic Estimation on Factor Graphs}

So we have a strong feeling that algorithms which can operate with
purely local computation and in-place data storage on a graph, and communicate via
message passing, will fit well with coming computer hardware. Let us
consider the computation involved in Spatial AI in more detail. As
laid out in~\cite{Davison:ARXIV2018}, Spatial AI problems inherently
involve
various graphs, and that paper made suggestions about the rough way in
which the storage and computation could be jointly arranged.
The key point, however, is that in Spatial AI there are various structured
sources of uncertain information (priors, cameras, other sensors)
from which, in real-time, we must
extract estimates of quantities (robot locations, map geometry, map labels, etc.) which are represented by variables which also have their own structure.

The fundamental theory for consistent fusion of
many uncertain sources of data is Bayesian probability
theory~\cite{Jaynes:Book2003}.
A very powerful and general representation of the probabilistic
structure of inference problems is the {\em factor graph}.
A factor graph is an undirected bipartite graph whose nodes are either variables
or factors. The variables are numerical parameters of a system whose values we wish to estimate, but are not directly
observable. The factors which join these variables represent constraints imposed by
measurements from sensors or other information about the system (such as priors)
which we are directly able to access.

Each factor is connected to the
subset of variables it depends on, and specifies the probabilistic
dependence of the observed measurement on the values of those variables.
A variable is denoted $x_i$, and a factor is
denoted $f_s$.  The subset of variables which is connected to a
particular factor $f_s$ is denoted $\vecx_s$. The interpretation is
that $f_s(\vecx_s) = p(z_s | \vecx_s)$   is the probability of the numerical measurement $z_s$ captured at node $s$
given the variables $\vecx_s$.

The bipartite connection pattern of a factor graph defines the
factorisability structure of the whole probabilistic model, in that
all factors $f_s$ are independent of each other. 
The vector
of all variables is $\vecx = (x_1, x_2, \ldots)^\top$, and therefore
the total joint probability distribution over all variables is the
product of all factors:
\beq
\label{eqn:factorprod}
p(\vecx) = \prod_s f_s(\vecx_s)
~.
\eeq
Interesting estimation problems in computer vision and robotics can
invariably be analysed to determine their factor graph structure, and
Dellaert and colleagues in particular \cite{Dellaert:Kaess:Foundations2017} have played a very
important role in increasing understanding of the power of the factor
graph  formulation in our field.

Now, as we will see, since each factor is a function of some subset of the variables,  this joint distribution is some tangled function of all of the variables involved in the graph.
Our goal in {\em inference} is to separate out one or more more variables
of interest (or all of them) and determine their {\em marginal}
probability distributions: individual probabilistic estimates which take all of the measurement information in the factors into account. 
The tangled form of
the product which is the full joint distribution means that this is
usually a computationally challenging problem.

The factor graph describing a Spatial AI problem may be very large and
complicated, and will continuously change due to live, asynchronous
measurements.
Practical inference methods therefore often make various
approximations, for instance by ignoring some measurements or priors,
or `baking in' certain aspects. For instance, in SemanticFusion (based
on ElasticFusion~\cite{Whelan:etal:RSS2015}), the 3D
reconstruction component runs by decoupling camera tracking and map
updates into independent, alternating estimation processes, with only
an occasional explicit loop closure optimisation to take account of
camera drift in an approximate way. Further, the semantic labelling
carried out is not used to improve the geometric estimation in any
way, though this would make a lot of sense (to apply map smoothing to
regions confidently labelled as floor or walls for instance). 

We know how to represent the ideal joint estimation problem to take
account of these measurements properly in a factor graph, but it has
not been feasible to do inference on such complicated graphs in
real-time in practical systems, due not only to to computational complexity 
but also system design complexity.
Again, approximate things could be
done such as pre-smoothing depth maps in response to semantic labelling before fusing them into
the 3D model, but this risks dangerous `double counting' of
information, which is known to lead to over-confident Bayesian estimates. The effect of a particular prior or
measurement should only appear once in the whole graph.

The purest representation of the knowledge in a Spatial AI problem is the factor graph itself, rather than probability distributions derived from it, which will always have to be stored with some approximation.
What we are really seeking is an algorithm which implements Spatial AI in a distributed way on a computational resource like a graph processor, by storing the factor graph as the master representation 
and operating on it in place using local computation and message
passing to implement estimation of variables as needed but taking
account of global influence. We imagine messages continually bubbling
around a large factor graph, which is changing continually with the
addition of new measurement factors and variable nodes, and perhaps
never reaching full convergence, but always being close in a way which
can be controlled. It may be that estimation processes will proceed in
an attention-driven way, using a lot of computation to bring high
quality to currently important areas or aspects, which then are
allowed to fade to a less up-to-date state once attention moves on, in
a `just-in-time' style 
\cite{Weerasekera:etal:ICRA2018}.

We will return to these general ideas in later discussion, but let us first get more concrete still about probabilistic estimation techniques.

\subsection{Distributable Estimation using Gaussian Distributions}

Almost all serious, scalable probabilistic estimation is based on the
core assumption of Gaussianity in `most' measurement distributions and
`most' posterior variable distrubutions, `most' of the time. We say this
with full knowledge that many other representations of distributions
have been used, from sampling to other explicit functional
parameterisations. But again and again, we come back to Gaussians due
to their fundamental properties of fitting real-world statistical
processes and the efficient representation of high-dimensional
distributions they allow as the `central' distribution of probability
theory~\cite{Jaynes:Book2003}.

The most important techniques in current geometric Spatial AI estimation are
all Gaussian-based techniques such as Extended Kalman Filtering and Bundle
Adjustment.  Gaussian-based methods have very close links to linear
algebra, because optimising Gaussian likelihoods is equivalent to
least-squares minimisation which involves the solution of linear
systems. When we write down the joint probability distribution (Equation~\ref{eqn:factorprod}) represented by a Gaussian factor graph, the result is a product of Gaussians. Finding the most probable variable values is equivalent to minimising the negative log of this probability distribution, which is a sum of terms which are quadratic functions of the variables. To find the minimum of this sum, we find the information matrix which depends of the Jacobians of the measurement functions with respect to the variables and the precision matrices of the measurements, and then must solve a linear system involving this information matrix. (This is done iteratively if the Gaussian measurement fuctions are non-linear in the variables.)

The key to the efficiency of this whole procedure is the form of the
information matrix (which has the same sparsity structure as  $\matJ^\top
\matJ$, often discussed in optimisation problems, as long as
measurement precision matrices are diagonal), which must be inverted to solve the linear system. 
Many decades of work have been devoted to studying the structure of
this matrix and efficient algorithms for inverting it. In Spatial AI,
extra interest and difficulty is due to the fact that the estimation
problem in question is incremental, with estimates needed in real-time
and measurements continually arriving. There has been much analysis of the
trade-offs between filtering approaches which marginalise out old
variables such as historic robot pose estimates and others which
repeatedly solve a whole estimation problem from
scratch~\cite{Strasdat:etal:ICRA2010}. 

Inverting large sparse information matrices has been tackled with a
variety of classic methods which take advantage of sparsity patterns
or some degree of parallelisation, such as 
Cholesky decomposition, Conjugate gradients, 
Jacobi, Gauss-Seidel, Red-Black ordering,
Multigrid, etc.
Good recent discussion of different optimisation methods in the context of
robot vision was given by PhD theses by Zienkiewicz~\cite{Zienkiewicz:PHD2017},
Engel~\cite{Engel:PhDThesis:2017} or Newcombe~\cite{Newcombe:PHD2012}.
Particular sub-problems in Spatial AI have well-known information
matrix structure. For instance, bundle adjustment for consistent scene
reconstruction, where a relatively
small number of cameras observe a large number of 3D points, has a
factor graph where every factor joins one pose variable to one point
variable, and on a CPU is well tackled using Cholesky
decomposition~\cite{Triggs:etal:VISALG1999}, or on a GPU by the
conjugate gradient method~\cite{Wu:etal:BA:CVPR2011}. Surface reconstruction on
a regular grid, where measurements from a sensor are combined with
smoothness priors, can be parallelised with methods like the Primal
Dual algorithm~\cite{Chambolle:2011}
Pure visual-inertial odometry can be tackled well with sliding window
filtering or non-linear
optimisation~\cite{Mourikis:Roumeliotis:ICRA2007,Leutenegger:etal:IJRR2014}. 

However, as discussed earlier, prototype
general Spatial AI systems need to have all of these elements and much
more, and highly-tuned specific estimation modules have often been
thrown together in unsatisfactory ways, requiring a lot of
approximation of probabilistic structure or heavy computing resources,
in particular with large data flows in and out of CPU RAM. General,
efficient and 
scalable Spatial AI estimation needs to cope with various different
dynamic factor graph patterns, involving priors and many types of
measurements flowing into the graph.
We need computation, storage and data
transfer characteristics well matched to both the modules and their
interfaces, and allowing practical incremental estimation.

Approaches such as iSAM2 from Kaess~\etal 
\cite{Kaess:etal:IJRR2012}
stand out as progress on taking a flexible approach to scalable
incremental estimation. iSAM2, a CPU algorithm, uses a dynamic data structure called the Bayes Tree
to represent a good approximation to the full factor graph of SLAM
problems in such as way that most updates can be carried out with
local message passing, with 
a more
substantial editing operation needed
only in response to rarer events such as loop closure.

We share the idea with iSAM2 of a factor graph as the master
representation, but with graph computing hardware in mind we believe
that we should be even more flexible.
If we wish estimation on factor graphs to have the properties of
purely local computation and data storage, we must get away
from the idea that a `god's eye view' of the whole structure of the
graph will ever be available.
We are guided towards methods where each node of a processing and
storage graph can operate with minimum knowledge of the whole graph
structure --- at a minimum, only purely local information about itself
and its near neighbours.

This is the character of belief propagation, which in its purest form
allows in-place inference on a factor graph with entirely local storage
and processing.
Each variable and factor node processes messages with no
knowledge about the rest of the graph other than its direct
neighbours, and BP can converge with arbitrary, asynchronous message
passing schedules which need no global coordination.
In a certain sense, an algorithm which works like this
represents `assuming the worst' --- that no knowledge of the structure
of an estimation problem is available to enable intelligent design of
processing.

This is the reason both that BP is well worth studying as an end point
in a continuum of possible methods, but also that it is unlikely to
form the whole solution to practical estimation. What we foresee is
that BP could form a general estimation `glue' between specifically
engineered hardware/software modules for particular tasks; or be
particularly valuable in highly dynamic, rapidly reconfiguring
estimation problems where management of computation can carry on in a
decentralised way.

We will show that Gaussian Belief Propagation is a general tool which
can be formulated for any standard problem that can be formulated as a
factor graph, and can for instance handle non-linear measurement
models and robust kernels.

Gaussian Belief Propagation has an extensive literature, and we are
not the first to consider applying it to vision, robotics and SLAM
problems, although we believe that it has received much less interest
than it deserves in this context. Non-Gaussian Belief Propagation is
much better known as a technique in computer vision for image
processing tasks on regular image grids.
Weiss and Freeman \cite{Weiss:Freeman:NIPS2000} did important work
showing the generality and correctness of loopy GBP in an AI context.

Most relevant to us, Ranganathan \etal in their `Loopy SAM' paper
\cite{Ranganathan:etal:IJCAI2007} showed GBP used for a robot mapping
application, and their experiments have many similarities with the
demonstrations we will give later in this paper.
We believe that despite the promising results in that paper, there was
not much follow-up due to the fact that the majority of researchers
have been concentrating on CPU performance.
Going back further, Paskin \etal \cite{Paskin:IJCAI2003} built a junction tree of a filtered SLAM graph which was kept sparse by removing edges and used GBP for inference. 
Work on Gaussian Processes in loopy graphs such by
Sudderth \etal~\cite{Sudderth:etal:TSP2004} is also related.
More recently, 
Crandall \etal  \cite{Crandall:etal:CVPR2011} used discrete BP to provide an initialisation for bundle adjustment, but then standard optimisation afterwards.

A more general research area which is strongly related is Multi-Robot SLAM~\cite{Saeedi:etal:JFR2016,Choudhary:etal:IJRR2017}, where many approaches to distributed mapping have been studied over the years, though usually not with the granularity of distribution that we are currently considering and often focusing on the assembly and sharing of a few discrete maps.

\section{Tutorial on Belief Propagation}

We will first introduce the general theory of Belief Propagation,
focusing on the Sum-Product Algorithm due to Pearl~\cite{Pearl:AAAI1982}, and 
following the notation and derivation given in Bishop's book `Pattern
Recognition and Machine Learning' \cite{Bishop:Book2006} (note that this excellent book has now been made available as a free download). Here the
representation of probability distributions is not specified,
and could be discrete probability tables or otherwise. We will go on to
derive the specific Gaussian case in Section~\ref{sec:gabp}, and
readers already familiar with BP could skip straight to that section.
We give a lot of detail on the mathematical derivations in these
sections, with the aim of making them fully understandable for the
committed reader.

We start from Equation~\ref{eqn:factorprod} which defines the probability distribution over all variables in a factor graph as a product of all factors, and remember that in inference we aim to determine the marginal distribution over variables of interest.
Choosing to start with one
particular variable $x$, its marginal distribution is found by taking the joint distribution, and
summing over all of the other variables:
\beq
\label{eqn:marg}
p(x) = \sum_{\vecx\setminus x} p(\vecx)
~,
\eeq
where the notation $\vecx\setminus x$ means all elements of $\vecx$
except $x$.

\begin{figure}[t]
\centerline{
\hfill
\includegraphics[width=\columnwidth]{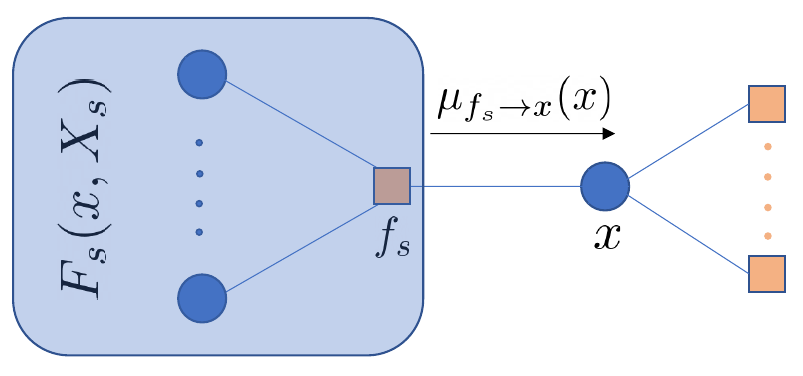}
\hfill
}
\caption{\label{fig:minitree}
A variable node $x$   connects to factor nodes including $f_s$, from
which it receives message $\mu_{f_s\rightarrow x}(x)$.
}
\vspace{2mm} \hrule
\end{figure}

For the moment, we will assume that our factor graph has a tree
structure, which means that it has no loops, and that there is
precisely one route through the graph between any two nodes.

Consider
Figure~\ref{fig:minitree} which focuses on an arbitrary variable $x$
within a tree factor graph. Variable $x$ is directly connected to a number of
factors $f_s$. Every other factor in the graph is
connected to $x$ indirectly via exactly one of these factors, so we
can divide the whole graph into the same number of subsets as the
factors $f_s$, and write the whole joint probability distribution as a
product of these subsets:
\beq
\label{eqn:Fs}
p(\vecx) = \prod_{s\in n(x)} F_s(x, \vecX_s)
~.
\eeq
Here $n(x)$ is the set of factor nodes that are neighbours of $x$;
$F_s$ is the product of all factors in the group associated with
$f_s$; and $\vecX_s$ is the vector of all variables in the subtree
connected to $x$ via $f_s$. Now, combining Equations~\ref{eqn:marg} and~\ref{eqn:Fs}:
\beq
\label{eqn:sumprod}
p(x) = \sum_{\vecx\setminus x} \left[ \prod_{s\in n(x)} F_s(x, \vecX_s) \right] ~.
\eeq
We can reorder the sum and product to obtain:
\beq
\label{eqn:prodsum}
p(x) = \prod_{s\in n(x)} \left[ \sum_{\vecX_s} F_s(x, \vecX_s) \right] ~.
\eeq
It is important to have a good intuition for what has happened with
this switch. Each term $F_s(x, \vecX_s)$ is the product of many factors; so
it is a multivariate function of $x$ and all of the other variables in
that branch of the tree. 
In Equation~\ref{eqn:sumprod}, we first multiply all of the $F_s$
terms together, to get a single joint function of all variables in the
whole tree. In the sum, we then marginalise out over all other
variables
to  be left with a marginal function only over our variable of
interest $x$.

In Equation~\ref{eqn:prodsum}, on the other hand, we perform
marginalisation first, taking each product of factors in a branch
$F_s(x, \vecX_s)$ and summing over all other variables to obtain a
function only of $x$ in the square bracket for each branch. We then just calculate the
product of these branch functions of $x$ to obtain the final marginal
distribution over $x$.

We can start to see now the idea of using message passing terminology
to describe this process. Continuing to use Bishop's notation, we
define:
\beq
\label{eqn:mufsx}
\mu_{f_s\rightarrow x}(x) = \sum_{\vecX_s} F_s(x, \vecX_s)
~.
\eeq
This term $\mu_{f_s\rightarrow x}(x)$ can be considered as a {\em
  message} from factor $f_s$ to variable $x$. The message has the form of a
probability distribution over variable $x$ only, and is the marginalised
probability over $x$ as the result of considering all factors in
one branch of the tree: it is `what that branch of the tree says about the marginal probability distribution of $x$'. If variable $x$ receives such a message from
all of the branches it is connected to, it can pool this information, and calculate its final
marginal distribution by simply multiplying these messages together:
\beq
\label{eqn:prodmes}
p(x) = \prod_{s\in n(x)} \mu_{f_s\rightarrow x}(x)
~.
\eeq

\begin{figure}[t]
\centerline{
\hfill
\includegraphics[width=\columnwidth]{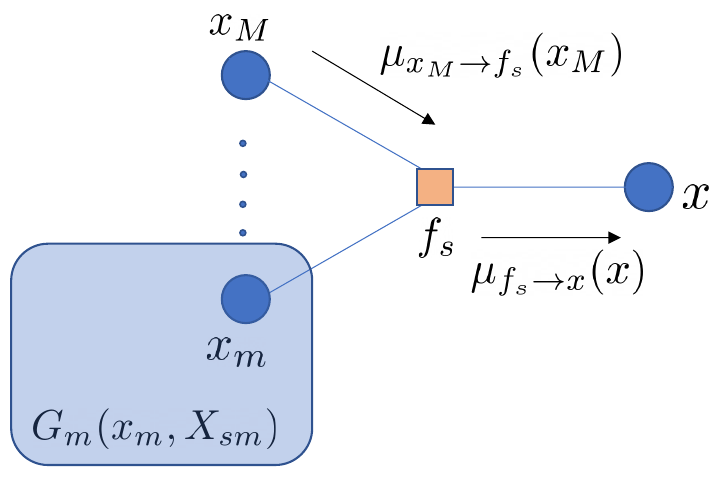}
\hfill
}
\caption{\label{fig:2messages}
Factor $f_s$ connects variable $x$ to $M$ other neighbouring variables $x_m \in x_1 \ldots x_M$, each of which is the root of a sub-branch containing a product of
factors $G_m$.
}
\vspace{2mm} \hrule
\end{figure}

\begin{figure}[t]
\centerline{
\hfill
\includegraphics[width=\columnwidth]{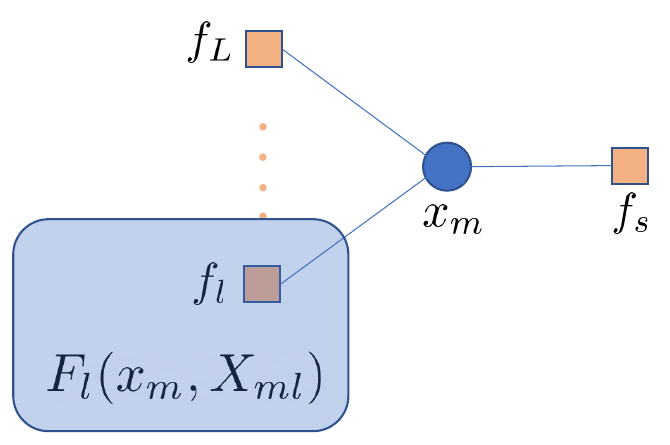}
\hfill
}
\caption{\label{fig:3factors}
$x_m$, one of the variable neighbours of $f_s$, connects $f_s$ to the product of factors $G_m(x_m, \vecX_{S_m})$ which we break down as $\prod_{l\in n(x_m) \setminus f_s} F_l(x_m,
\vecX_{m_l})$.
}
\vspace{2mm} \hrule
\end{figure}

Next, we go further into one of the branches of the tree, and break down the products of factors $F_s(x, \vecX_s)$ as follows:
\bea
F_s(x, \vecX_s) &=& f_s(x, x_1, \ldots, x_M) \\
&& \times
G_1(x_1, \vecX_{S_1}), \ldots, G_M(x_M, \vecX_{S_M}) \nonumber\\
&=& f_s(x, x_1, \ldots, x_M) \prod_{m\in n(f_s)} G_m(x_m, \vecX_{S_m})
\nonumber\\
&&
\eea
Here, referring to Figure~\ref{fig:2messages}, $f_s$, the factor which connects $x$ to this branch, is a function of $x$ as well as
$M$ other neighbouring variables $x_m \in x_1 \ldots x_M$. Each of
these variables connects to a sub-branch containing a product of
factors $G_m$
which is a function of variable $x_m$ and other variables $\vecX_{S_m}$.
Substituting into Equation~\ref{eqn:mufsx}:
\bea
\label{eqn:mufsx2}
& \mu_{f_s\rightarrow x} & (x) \\
&=& \sum_{\vecX_s} \left[ f_s(x, x_1, \ldots, x_M) \prod_{m\in n(f_s)} G_m(x_m, \vecX_{S_m}) \right] \nonumber \\
&=& \sum_{x_1, \ldots, x_M} \sum_{\vecX_{S_1}, \ldots, \vecX_{S_M}} \\
&&
\left[ f_s(x, x_1, \ldots, x_M)
 \prod_{m\in n(f_s)} G_m(x_m, \vecX_{S_m}) \right] \nonumber \\
&=& \sum_{x_1, \ldots, x_M} f_s(x, x_1, \ldots, x_M) \\
&&
\prod_{m\in n(f_s)} \sum_{\vecX_{S_1}, \ldots, \vecX_{S_M}} G_m(x_m, \vecX_{S_m}) ~,
\eea
where we have made use of the fact that $\vecX_s =   (x_1, \ldots, x_M, \vecX_{S_1}, \ldots, \vecX_{S_M})$ to separate out the sum.
We can now define the second type of message, this time from variable to factor:
\beq
\label{eqn:vartofac}
\mu_{x_m\rightarrow f_s}(x_m) = \sum_{\vecX_{s_m}} G_m(x_m, \vecX_{S_m})
~,
\eeq
and substitute into Equation~\ref{eqn:mufsx2} to get:
\beq
\label{eqn:mufsx3}
\mu_{f_s\rightarrow x}(x) = \sum_{x_1, \ldots, x_M} f_s(x, x_1,
\ldots, x_M) \prod_{m\in n(f_s)} \mu_{x_m\rightarrow f_s}(x_m)
~.
\eeq
We see here one half of the full recursive solution we are looking
for: an expression for messages from factors to variables in terms of
the messages those factors have received from other variables. We need
just a few more steps to find ther other half of this.
We need to take one more step deeper into the tree. Consider Figure~\ref{fig:3factors}, which now centres on $x_m$, one of the variable neighbours of $f_s$, which connects $f_s$ to the product of factors $G_m(x_m, \vecX_{S_m})$. We break down this product as follows:
\beq
G_m(x_m, \vecX_{S_m}) = \prod_{l\in n(x_m) \setminus f_s} F_l(x_m,
\vecX_{m_l})
  ~.
\eeq
We see that the total product factorises into terms
$F_l(x_m,\vecX_{m_l})$, each of which is the product of the set of
factors from the whole graph which connects to $x_m$ via factor
$f_l$. (We have broken down $\vecX_{S_m}$, the set of all variables
connected to $f_s$ via $x_m$, into subsets $\vecX_{m_l}$ which connect
to $x_m$ via factor $f_l$.)

We substitute this factorisation into Equation~\ref{eqn:vartofac}:
\beq
\mu_{x_m\rightarrow f_s }(x_m) = \sum_{\vecX_{s_m}}
\prod_{l\in n(x_m) \setminus f_s} F_l(x_m,
\vecX_{m_l})
~,
\eeq
and as we have seen before swap the order of the sum and product to
obtain:
\beq
\mu_{x_m\rightarrow f_s}(x_m) = 
\prod_{l\in n(x_m) \setminus f_s}
\sum_{\vecX_{m_l}}
F_l(x_m,
\vecX_{m_l})
~.
\eeq
Here we recognise the form of a message from factor to variable as
defined in Equation~\ref{eqn:mufsx}, and substitute to obtain:
\beq
\label{eqn:muxfsrec}
\mu_{x_m\rightarrow f_s}(x_m) = 
\prod_{l\in n(x_m) \setminus f_s}
\mu_{f_l\rightarrow x_m}(x_m)
~.
\eeq
We now have all we need for the full sum-product algorithm, and can
focus on Equations~\ref{eqn:prodmes}, \ref{eqn:mufsx3} and
\ref{eqn:muxfsrec}. Equation~\ref{eqn:prodmes} says that in order to
calculate the marginal distribution for $x$, we should multiply
together all of messages received from each of its neighbouring factor
nodes. Each of those messages has the form of a probability
distribution over $x$ only.

Stepping out to any one of the neighbouring factor nodes, we see the work that needs to be done at a factor node in Equation~\ref{eqn:mufsx3}. A factor node receives messages from a number of variables, and must calculate a new message to send out.
The messages that factor has received from other variables, each of which is a function of that one other variable, are all multiplied.   We then multiply this product by the probability distribution representing the factor itself. We then marginalise out all variables other than the one to which the message will be sent, to leave a function of that variable only and that is the message that is sent.

One more step out,  Equation~\ref{eqn:muxfsrec}  shows what happens at a
variable node. It receives messages from a number of factors, all of
which are functions of the variable, and multiplies these together to
generate the message it sends on to the next factor.

It should now be clear that these two steps are simply repeated
recursively through the whole tree. In order to find the marginal
distribution for $x$, we start from all of the leaf nodes of the
factor graph relative to $x$, which can be either variables or nodes,
and pass messages inwards towards $x$. When each node has received
messages from all outer nodes, it can perform its calculation to
generate the correct message to pass inwards. This continues recursively all the way to $x$ at the root of the tree.

One remaining detail is how to initialise the leaf nodes, and this is
simply dealt with. A variable leaf node sends a message $\mu_{x
  \rightarrow f}(x) = 1$ to its only connected factor, and a factor
leaf node sends $\mu_{f \rightarrow x}(x) = f(x)$.  These are seen to
be correct from looking at Equations 
\ref{eqn:mufsx3} and
\ref{eqn:muxfsrec} if we imagine a set of null factors with flat
probability distributions surrounding the main tree.

So we know how to find the marginal distribution at a chosen node $x$
within a tree by defining that node as the root and passing messages
recursively in towards it from all of the leaf nodes. If we required
marginal distributions for {\em all} variable nodes in the tree,
clearly we could simply repeat this procedure for each one. However,
this would require a huge amount of wasted work. Imagine two variable
nodes which are close together in a large tree. Defining either as the
root node would lead to large equal branch and leaf structures in
distant  parts of the tree, and exactly the same computation in these
regions would be repeated.

In fact, it is quite easy to see that we can find the marginal
distribution for every variable node using only double the amount of
work required to find the marginal for one variable. During the leaves-to-root message passing procedure to find the marginal for $x$, every
variable and factor node along the way will have received incoming
messages from all of its neighbours apart from the one to which it
must transmit an outgoing message in the direction towards $x$. Once
the messages get all the way the $x$, the root is then `fully
informed' and has a final marginal distribution which takes into
account all of the measurement information in the graph. Therefore, if
we now send a second series of messages outwards from the root back to
the leaves, we will fill in the missing incoming message for every
variable node and can therefore calculate a fully informed marginal
for each one.

So belief propagation is able to efficiently determine marginal
distributions for every variable in a tree graph with a one time
forward/backward sweep of message passing through the graph. Most
factor graphs for practical estimation problems are not trees however,
but contain loops. This leads to two possibilities for the use of BP
methods. One is to convert  a general graph into a tree by combining
nodes via graph cliques. These will be perfect trees, but with large
compound nodes, and leads to the junction tree family of methods.
The other is to retain the full loopy graph, but apply BP methods as
if the graph was a tree, and keep iterating until convergence is
reached. This approach is called {\em loopy belief propagation}, and
has been shown to converge to useful solutions in many problems. We
will test this out later, but first go on to derive the theory for
belief propagation in the specific case that all probability
distributions are Gaussian.

\section{Gaussian Belief Propagation}
\label{sec:gabp}

In the case where the relationship between factors and variables is
linear, and all factors have a Gaussian probability distribution, it
is well understood that inference leads to a multivariate Gaussian
distribution over the variables.
It is also very well established that factor graphs which have factors
which are Gaussians with non-linear dependence on the variables can be
solved using efficient second order iterative optimisation (this is
the class of non-linear least squares methods).

Here we will show how belief propagation can be implemented in this
Gaussian case which is the norm in robotics and computer vision. Note
that we have switched from scalars to vectors here for variable
state space.

\subsection{Factor Definition}

We will start with a general specification of a factor which will be
familiar to anyone used to probabilistic estimation in computer vision
and robotics.
Suppose that a robot has a sensor which is configured to observe a
quantity which is a function of the state variables of the robot. When
tested, the sensor is found to report measurements which differ from
the expected `ground truth' in a way described by a Gaussian distribution.
We define the associated Gaussian factor as follows:
\beq
\label{equ:generalfactor}
f_s(\vecx_s) = K e^{-\frac{1}{2} \left[ (\vecz_s - \vech_s(\vecx_s))^\top \matLambda_s (\vecz_s - \vech_s(\vecx_s))   \right]  }
~.
\eeq
This expression represents the probability of obtaining vector
measurement $\vecz_s$ from the sensor.
Factor $f_s$ is a function of the set of involved variables 
$\vecx_s$, a subset of the whole state $\vecx$.
The form of the function is a squared exponential with
a scaling factor $K$ for normalisation whose value we will not need to
calculate. Within the exponential, we see an inner product. This involves $\vech_s$, the
function which describes the dependence of the measurement on the
variables, and $\vecz_s$, the value actually measured. Matrix
$\matLambda_s$ is the precision or inverse covariance of the measurement.

Note that we can also use factors of this form for priors which are
not sensor measurements but assumptions or external knowledge, such as
smoothness priors. We will look at the details of this in our examples later.

In summary, to specify a Gaussian factor, we need:
\bi
\item $\vech_s(\vecx_s)$, the functional form of the dependence of the measurement on the local state variables.
\item $\vecz_s$, the actual observed value of the measurement.
\item $\matLambda_s$, the symmetric precision matrix of the measurement.
\ei

\subsection{State Representation}

In GBP, the probability
distributions over state variables also have a Gaussian form. A Gaussian distribution in the
state space of a particular variable node $m$ is generally written as:
\beq
\label{eqn:covform}
p_m(\vecx_m) = K e^{-\frac{1}{2} \left[ (\vecx_m - \vecmu_m)^\top \matLambda_m (\vecx_m - \vecmu_m)   \right]  }
~,
\eeq
where $\vecmu_m$ is the mean of the distribution and $\matLambda_m$ is
its precision or inverse covariance. An equivalent alternative form
is:
\beq
\label{eqn:infoform}
p_m(\vecx_m) = K_2 e^{ \left[ -\frac{1}{2} \vecx_m^\top \matLambda_m \vecx_m
+ \veceta_m^\top \vecx_m  \right]  }
~.
\eeq
This is the information form, as explained very clearly for SLAM readers by Eustice~\etal~\cite{Eustice:etal:ICRA2005}. Note the different constant factor $K_2 \neq K$, but we will not need to calculate the value of either.
The information vector $\veceta_m$
is related to the mean vector by the relation:
\beq
\veceta_m = \matLambda_m \vecmu_m
~.
\eeq
From here on, we will represent a Gaussian distribution over vector
$\vecx_m$ in this information form, using vector $\veceta_m$ and
precision matrix $\matLambda_m$. The information form is preferred to the covariance form as it can represent rank deficient Gaussian distributions with zero information which would correspond to an infinite covariance along dimensions which are fully unconstrained. The information form is also convenient as multiplication of distributions is handled simply by adding information vectors and precision matrices.

\subsection{Linearising Factors}

As in all types of scalable Gaussian-based estimation, we need to be able to produce a local linear version of any general non-linear factor in the form of Equation~\ref{equ:generalfactor}. Following Equations~\ref{eqn:covform} and~\ref{eqn:infoform}, a linear factor has the form of a Gaussian distribution over the variables $\vecx_s$ involved in the factor, expressed either in mean/precision form as:
\beq
f_s(\vecx_s) = K e^{-\frac{1}{2} \left[ (\vecx_s - \vecmu_s)^\top \matLambda_s' (\vecx_s - \vecmu_s)   \right]  }
~,
\eeq
or information form: 
\beq
\label{eqn:coninf}
f_s(\vecx_s) = K_2 e^{\left[-\frac{1}{2} \vecx_s^\top \matLambda_s' \vecx_s + \veceta_s^\top \vecx_s \right]    }
~.
\eeq
where mean $\vecmu_s$, precision $\matLambda_s'$ and information vector $\veceta_s$ are related by:
\beq
\veceta_s = \matLambda_s' \vecmu_s
~.
\eeq
Concentrating on the information form of Equation~\ref{eqn:coninf}, we
need to find the values of $\veceta_s$ and $\matLambda_s'$ to linearise
the nonlinear constraint of Equation~\ref{equ:generalfactor} around
a state estimate $\vecx_s = \vecx_0$.
First, let us rewrite Equation~\ref{equ:generalfactor} as:
\beq
\label{eqn:en}
f_s(\vecx_s) = K e^{-\frac{1}{2} E_s}
~,
\eeq
where $E_s$, the least squares `energy' of the constraint, is:
\beq
\label{eqn:energy}
E_s = (\vecz_s - \vech_s(\vecx_s))^\top \matLambda_s (\vecz_s -
\vech_s(\vecx_s))
~.
\eeq
Then we apply the first order Taylor series expansion of non-linear
measurement function $\vech_s$ to find its approximate value for state
values $\vecx_s$ close to $\vecx_0$:
\beq
\vech_s(\vecx_s) \approx \vech_s(\vecx_0) + \matJ_s (\vecx_s -
\vecx_0)
~,
\eeq
where $\matJ_s$ is the Jacobian $\frac{\partial \vech_s}{\partial
  \vecx_s}\big|_{\vecx_s = \vecx_0}$.
Substituting into Equation~\ref{eqn:energy} and rearranging:
\bea
E_s &=& \left[\vecz_s - (\vech_s(\vecx_0) + \matJ_s (\vecx_s -
\vecx_0))   \right]^\top \nonumber \\ &&\matLambda_s \left[\vecz_s -
(\vech_s(\vecx_0) + \matJ_s (\vecx_s -
\vecx_0))   \right] 
\nonumber \\
&=& \left[(\vecz_s - \vech_s(\vecx_0))  - \matJ_s (\vecx_s -
  \vecx_0))   \right]^\top \nonumber \\ &&
\matLambda_s
\left[(\vecz_s - \vech_s(\vecx_0))  - \matJ_s (\vecx_s -
  \vecx_0))   \right]
\nonumber \\
&=&
(\vecz_s - \vech_s(\vecx_0))^\top 
\matLambda_s
(\vecz_s - \vech_s(\vecx_0)) \nonumber \\
&&  -
(\vecz_s - \vech_s(\vecx_0))^\top 
\matLambda_s
  \matJ_s (\vecx_s -
  \vecx_0) \nonumber \\
&&  -
  (\matJ_s (\vecx_s -
  \vecx_0))^\top
  \matLambda_s
  (\vecz_s - \vech_s(\vecx_0))
    \nonumber \\
&&  +
   (\matJ_s (\vecx_s -
  \vecx_0))^\top
  \matLambda_s
  \matJ_s (\vecx_s -
  \vecx_0)
  ~.
\eea
The first of the four terms here is a constant which doesn't depend on
$\vecx_s$, and the second and third are equal (one is the transpose of
the other, and both are scalars), so we can simplify to:
\bea
E_s &=& K_3
- 2
(\vecz_s - \vech_s(\vecx_0))^\top 
\matLambda_s
  \matJ_s (\vecx_s -
  \vecx_0) \nonumber \\
&&+ 
   (\matJ_s (\vecx_s -
  \vecx_0))^\top
  \matLambda_s
  \matJ_s (\vecx_s -
  \vecx_0)
  \nonumber \\
  &=&
  (\vecx_s -
  \vecx_0)^\top \matLambda_s' (\vecx_s -
  \vecx_0) \nonumber \\
&&- 2
(\vecz_s - \vech_s(\vecx_0))^\top 
\matLambda_s
  \matJ_s (\vecx_s -
  \vecx_0) + K_3\nonumber
  ~,
\eea
where $\matLambda_s' = \matJ_s^\top \matLambda_s \matJ_s$. Going further:
\bea
E_s &=&
\vecx_s^\top \matLambda_s' \vecx_s
- \vecx_0^\top \matLambda_s' \vecx_s
- \vecx_s^\top \matLambda_s' \vecx_0
+ \vecx_0^\top \matLambda_s' \vecx_0 \nonumber\\
&&
- 2 (\vecz_s - \vech_s(\vecx_0))^\top 
\matLambda_s \matJ_s \vecx_s \nonumber \\
&&+ 2 (\vecz_s - \vech_s(\vecx_0))^\top 
\matLambda_s \matJ_s \vecx_0
+ K_3
~.
\eea
Here the second and third terms are equal, and the fourth and sixth
are constant, so:
\bea
E_s &=&
\vecx_s^\top \matLambda_s' \vecx_s 
- 2 \vecx_0^\top \matLambda_s' \vecx_s \nonumber \\
&&- 2 (\vecz_s - \vech_s(\vecx_0))^\top 
\matLambda_s \matJ_s \vecx_s 
+ K_4 \label{eqn:nearly}
~.
\eea
From Equations~\ref{eqn:coninf} and~\ref{eqn:en} we see that the
least squares energy in a linear constraint in the information form
is:
\beq
E_s = \vecx_s^\top \matLambda_s' \vecx_s - 2 \veceta_s^\top \vecx_s
~.
\eeq
Matching this with Equation~\ref{eqn:nearly}:
\beq
\veceta_s^\top = \vecx_0^\top \matLambda_s' + (\vecz_s -
\vech_s(\vecx_0))^\top \matLambda_s \matJ_s
~.
\eeq
And therefore, finally:
\bea
\veceta_s &=&
\matLambda_s' \vecx_o
+ \matJ_s^\top \matLambda_s(\vecz_s - \vech_s(\vecx_0)) \nonumber\\
&=& \matJ_s^\top \matLambda_s \left[
\matJ_s \vecx_0 + \vecz_s - \vech_s(\vecx_0)
\right]
~.
\eea
To summarise, to linearise a general non-linear factor $\vech_s(\vecx_s)$ around state variables $\vecx_0$, turning it into a Gaussian factor expressed in terms of $\vecx_s$,
we use the linear factor represented in information form by
information vector $\veceta_s$ and precision matrix $\matLambda_s'$
calculated as follows:
\bea
\veceta_s &=& \matJ_s^\top \matLambda_s \left[
  \matJ_s \vecx_0 + \vecz_s - \vech_s(\vecx_0) \right] \nonumber \\
\matLambda_s' &=& \matJ_s^\top \matLambda_s \matJ_s
\label{eqn:lin}
~.
\eea

\subsection{Message Passing at a Variable Node}

Let us now consider the computation which happens at nodes to
implement message passing. 
Remember that in Belief Propagation, messages always have the form of
a probability distribution in the state space of the variable node
either sending or receiving the message. In GBP, each message will
therefore take the form of an information vector and precision matrix
in that state space.

First, we consider the processing that happens at a variable node $\vecx_m$
during message passing. A variable node is connected to a number of
factors, and during a typical message passing step it receives
incoming messages from all of these except one, and must generate an
outgoing message to send to the remaining factor.
Here we follow the recipe of Equation~\ref{eqn:muxfsrec}. All of the
messages involved are in the state space of node $\vecx_m$. We simply need
to multiply together all of the incoming messages to generate the
outgoing message.

Each incoming message $\mu_{f_l\rightarrow \vecx_m}(\vecx_m)$ is represented
by an information vector $\veceta_{ml}$ and a precision matrix
$\matLambda_{ml}$. We obtain the information vector and precision
matrix of the outgoing message $\mu_{\vecx_m\rightarrow f_s}(\vecx_m)$ by
simply adding:
\bea
\veceta_{ms} &=& \sum_{l\in n(\vecx_m) \setminus f_s} \veceta_{ml}\\
\matLambda_{ms} &=& \sum_{l\in n(\vecx_m) \setminus f_s} \matLambda_{ml}
~.
\eea
This is because when we multiply several Gaussian expressions
of the form in Equation~\ref{eqn:infoform}, we add the exponents.

\subsection{Message Passing at a Factor Node}

A factor node receives incoming messages from a number of variable
nodes, and must process these to produce an outgoing message to the
target variable node. Following 
Equation~\ref{eqn:mufsx3}, the incoming messages are multiplied
together, and this product is also multiplied by the factor
distribution itself. Now each of the incoming messages is a function
of the state space of the variable node it comes from, while the
factor potential is a function of all of the variables connected to
the factor, including the output variable. The full product is
therefore also a function of all variables. Finally, all variables
other than the output variable are marginalised out from this joint
distribution, and the result is a function only in the output
variable's state space, and this is the outgoing message.

The 
general non-linear factor
 $\vech_s(\vecx_s)$ must be first linearised to an information vector $\veceta_s$
and precision matrix $\matLambda_s'$ as in
Equation~\ref{eqn:lin}. This linearisation requires anchor values
$\vecx_0$ of all of the connected variables, including the output
variable. This can be done once every message passing step, or much
less frequently. Clearly if a factor is a linear function in the first
place then we formulate the linear constraint once and do not
need to change it.

To the information vector and precision matrix representing the
constraint, we add the incoming messages from input variables.
This vector and matrix addition is done `in place'. In the vector of
variables $\vecx_s$ associated with the factor, there should be a partitioning
into contiguous sets which come from each connected variable
node. E.g. let us consider a factor with three connected variable
nodes, where $m_1$
$m_2$ are input nodes and $m_3$ is the output node in this case.
In the factor definition:
\beq
\vecx_s = \vecthree{\vecx_{sm_1}}{\vecx_{sm_2}}{\vecx_{sm_3}}
~.
\eeq
The information vector and precision matrix are partitioned in the
same way:
\bea
\veceta_s &=& \vecthree{\veceta_{sm_1}}{\veceta_{sm_2}}{\veceta_{sm_3}} \\
\matLambda_s' &=& \matthree{\matLambda_{sm_1m_1}'}{\matLambda_{sm_1m_2}'}{\matLambda_{sm_1m_3}'}{\matLambda_{sm_2m_1}'}{\matLambda_{sm_2m_2}'}{\matLambda_{sm_2m_3}'}{\matLambda_{sm_3m_1}'}{\matLambda_{sm_3m_2}'}{\matLambda_{sm_3m_3}'}
~.
\eea
So when conditioning on the messages coming from input notes $m_1$ and $m_2$ we get:
\beq
\label{eqn:etacond}
\veceta_{Cs} = \vecthree{\veceta_{sm_1} + \veceta_{m_1s}}{\veceta_{sm_2} + \veceta_{m_2s}}{\veceta_{sm_3}} \\
\eeq
\beq
\label{eqn:lambdacond}
\matLambda_{Cs}' = \matthree{\matLambda_{sm_1m_1}' + \matLambda_{m_1s}}{\matLambda_{sm_1m_2}'}{\matLambda_{sm_1m_3}'}{\matLambda_{sm_2m_1}'}{\matLambda_{sm_2m_2}'+ \matLambda_{m_2s}}{\matLambda_{sm_2m_3}'}{\matLambda_{sm_3m_1}'}{\matLambda_{sm_3m_2}'}{\matLambda_{sm_3m_3}'}
~.
\eeq
To complete message passing, from this joint distribution we must marginalise out all variables but those of the output node, in this example $m_3$. Eustice~\etal~\cite{Eustice:etal:ICRA2005} give the formula for marginalising a general partioned Gaussian state in information form. If the joint distribution is:
\bea
\label{eqn:alpha}
\veceta &=& \vectwo{\veceta_\alpha}{\veceta_{\beta}}\\
\label{eqn:beta}
\matLambda &=& \mattwo{\matLambda_{\alpha\alpha}}{\matLambda_{\alpha\beta}}{\matLambda_{\beta\alpha}}{\matLambda_{\beta\beta}}
~,
\eea
then marginalising out the $\beta$ variables to leave a distribution only over the $\alpha$ set is achieved by:
\bea
\label{eqn:emarg}
\veceta_{M\alpha} &=& \veceta_\alpha - \matLambda_{\alpha\beta} \matLambda_{\beta\beta}^{-1} \veceta_\beta \\
\label{eqn:Lmarg}
\matLambda_{M\alpha} &=& \matLambda_{\alpha\alpha} - \matLambda_{\alpha\beta} \matLambda_{\beta\beta}^{-1} \matLambda_{\beta\alpha} 
~.
\eea

To apply these formulae to the partitioned state of Equations~\ref{eqn:etacond} and~\ref{eqn:lambdacond}, we first reorder the vector and matrix to bring the output variable to the top. For our example where $m_3$ is the output variable, we reorder the conditioned vector and matrix:
\bea
\veceta_{Cs} &=& \vecthree{\veceta_{{Cs}m_1}}{\veceta_{{Cs}m_2}}{\veceta_{{Cs}m_3}} \\
\matLambda_{Cs}' &=& \matthree{\matLambda_{{Cs}m_1m_1}'}{\matLambda_{{Cs}m_1m_2}'}{\matLambda_{{Cs}m_1m_3}'}{\matLambda_{{Cs}m_2m_1}'}{\matLambda_{{Cs}m_2m_2}'}{\matLambda_{{Cs}m_2m_3}'}{\matLambda_{{Cs}m_3m_1}'}{\matLambda_{{Cs}m_3m_2}'}{\matLambda_{{Cs}m_3m_3}'}
\eea
to:
\bea
\label{eqn:eCR}
\veceta_{CRs} &=& \vecthree{\veceta_{{Cs}m_3}}{\veceta_{{Cs}m_1}}{\veceta_{{Cs}m_2}} \\
\label{eqn:LCR}
\matLambda_{CRs}' &=& \matthree
          {\matLambda_{{Cs}m_3m_3}'}{\matLambda_{{Cs}m_3m_1}'}{\matLambda_{{Cs}m_3m_2}'}
          {\matLambda_{{Cs}m_1m_3}'}{\matLambda_{{Cs}m_1m_1}'}{\matLambda_{{Cs}m_1m_2}'}
          {\matLambda_{{Cs}m_2m_3}'}{\matLambda_{{Cs}m_2m_1}'}{\matLambda_{{Cs}m_2m_2}'}
\eea
We then identify subblocks $\alpha = m_3$ and $\beta = m_1m_2$ between
Equations~\ref{eqn:alpha}, \ref{eqn:beta} and Equations~\ref{eqn:eCR},
\ref{eqn:LCR}, and apply Equations~\ref{eqn:emarg} and~\ref{eqn:Lmarg} to obtain the marginalised distribution over $m_3$ which forms the outgoing message $\mu_{f_s\rightarrow\vecx_{m_3}}$ to variable node $m_3$.

\subsection{Implemention Details}

One of the great strengths of GBP is the  straightforward and fully
local nature  of
implementation. We believe that  whereas previous estimation methods
are instantiated in  large and highly optimised `solver' libraries for
a CPU, the details of GBP can be easily and efficiently implemented
on any particular distributed platform, and what is more likely to
emerge is a set of standard formats for how these platforms should
pass messages between them, befitting our prediction of great value
for GBP as the glue between other estimation methods.

Our current simple CPU implementation is a prototype for future implementation on a graph processor or other distributed device, and therefore has decentralisation of data and processing into classes.
A VariableNode or FactorNode class object is instantiated for each
node of the factor graph. Specialisations of these implement the
particular state space and factor function models of the graph in
question, and the algorithms for message passing. However, these
classes do not store any state information. An Edge object is
instantiated for every connection between a variable and a factor, and
stores the latest VariableToFactorMessage and FactorToVariableMessage
for this link. So when a VariableNode or FactorNode needs to carry out
a message passing step, it reads the appropriate incoming messages
from all Edges it is connected to apart from one, performs the
calculation, and then writes its outgoing message to the other Edge. 

If we wish to form a best up-to-date estimate at a VariableNode at any
point in time, for
instance for visualisation, we can read and add FactorToVariableMessages from
{\em all} connected Edges.
Similarly, if a FactorNode reads and adds all incoming
VariableToFactorMessages to its factor potential  at any moment, we
get the current estimate of its energy based on all information
available, which can be used for instance to relinearise it (or as we
will see later, to apply a robust weighting).

In the limit of a purely distributed implementation, each node (either variable or
factor) could be hosted on separate processor, or tile of a graph processor.
The most intensive computation a node needs to carry out is the matrix
inversion needed for marginalisation at a factor node
($\matLambda_{\beta\beta}^{-1}$ for use in Equations~\ref{eqn:emarg}
and~\ref{eqn:Lmarg}). The dimension of this matrix is usually small. In
the common case of graphs with only unary or binary factors (which
connect to one or two variable nodes), the maximum dimension of
$\beta$ is the maximum individual variable node dimension.

As we will see in our demonstrations, we have found the convergence
of GBP to be remarkably independent of the ordering of message
passing schedules, and this is very promising for wide adoption
particularly in cases of multiple independent devices.  Initialisation
is usually also not problematic, because our parameteristion of
Gaussian distributions in the information form means that we can
safely represent the uncertainty over variables even if the factors
connecting to them do not fully constrain their degrees of freedom
(i.e. the precision matrices are singular) and covariances would not
be defined. However, if we do wish to visualise uncertainties from a
covariance we can add weak stabilising unary factors.

\section{Examples}

Before discussing more general issues, we now give examples of
simple implementations of GBP in settings relevant to Spatial AI.

\subsection{1D Surface Reconstruction}

\label{section:1d}

\begin{figure}[t]
	\centerline{
		\hfill
		\includegraphics[width=\columnwidth]{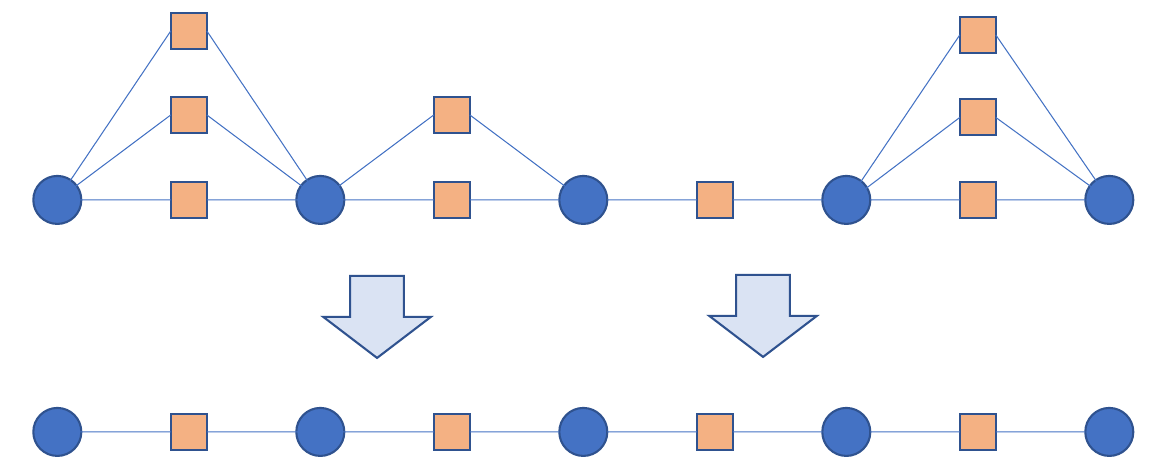}
		\hfill
	}
	\caption{\label{fig:surfacegraph}
		Factor graph for 1D surface reconstruction. Blue variable nodes are evenly spaced along the surface. Every consecutive pair is linked by a smoothness factor (orange, in line with the variable nodes); and where available, also by one or more measurement factors. Since all factors are linear, we combine all factors between a pair of variables into a single compound factor, to give the purely linear factor graph shown at the bottom. 
	}
	\vspace{2mm} \hrule
\end{figure}

\begin{figure*}
\centerline{
\hfill
\includegraphics[width=0.245\linewidth]{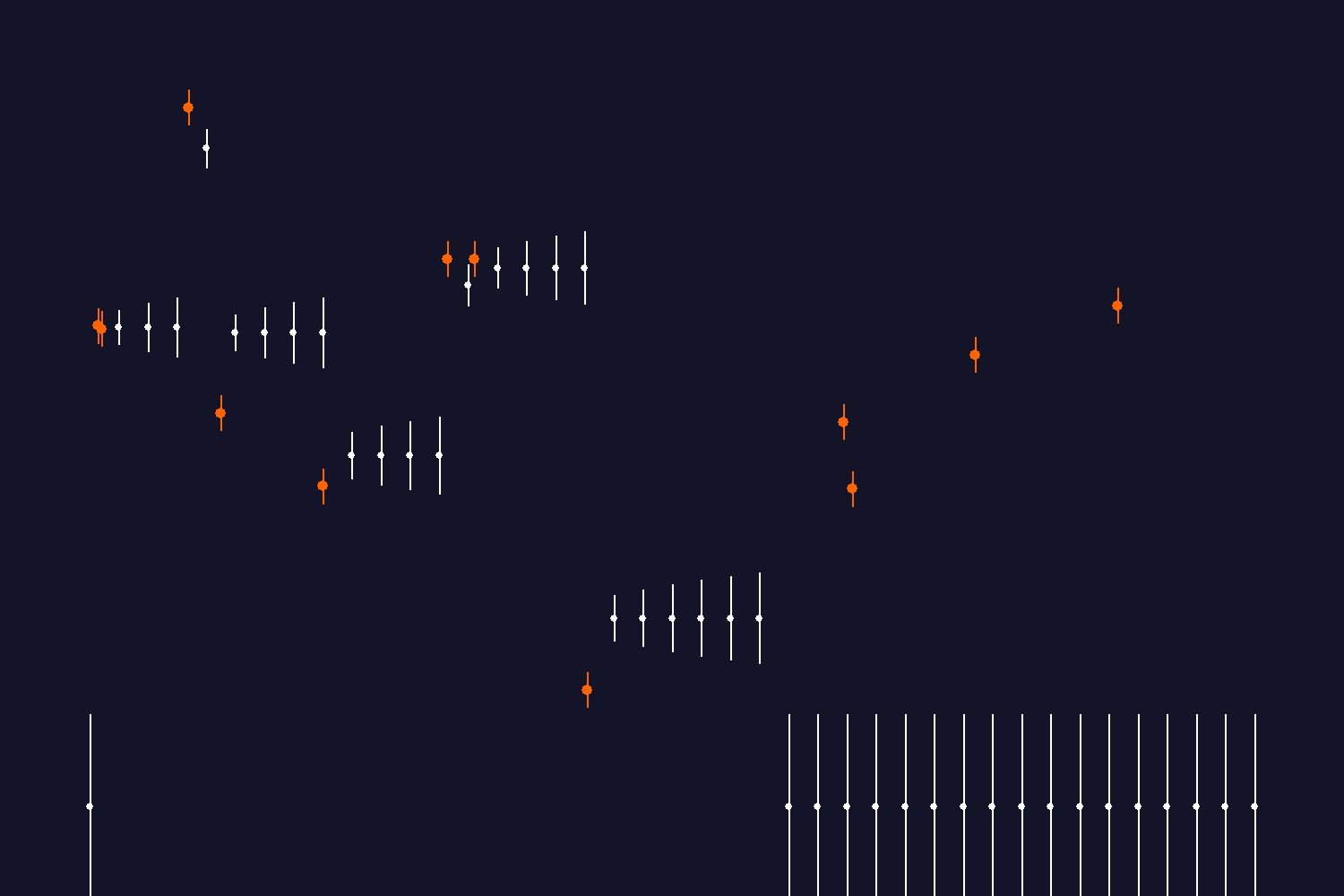}
\hfill
\includegraphics[width=0.245\linewidth]{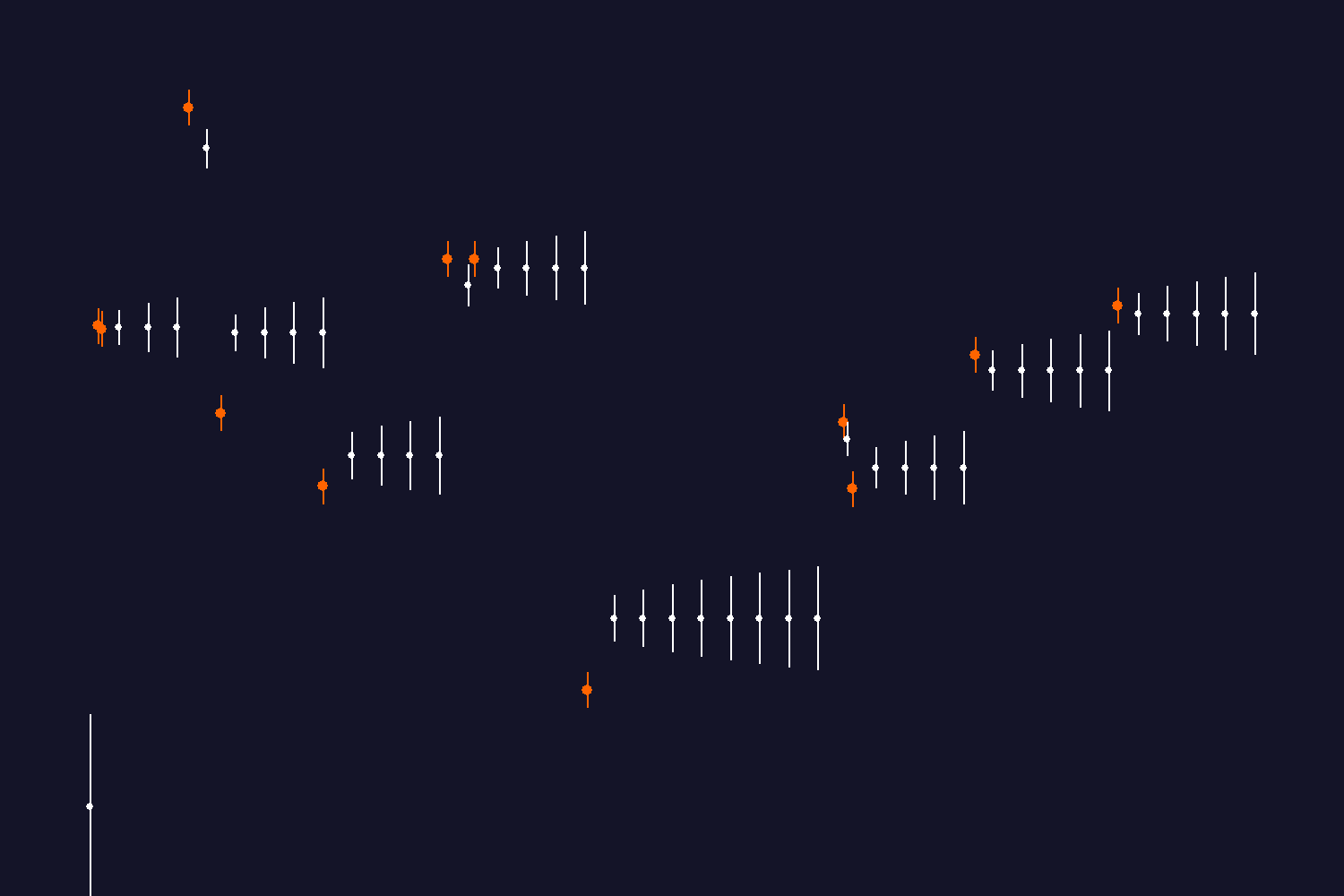}
\hfill
\includegraphics[width=0.245\linewidth]{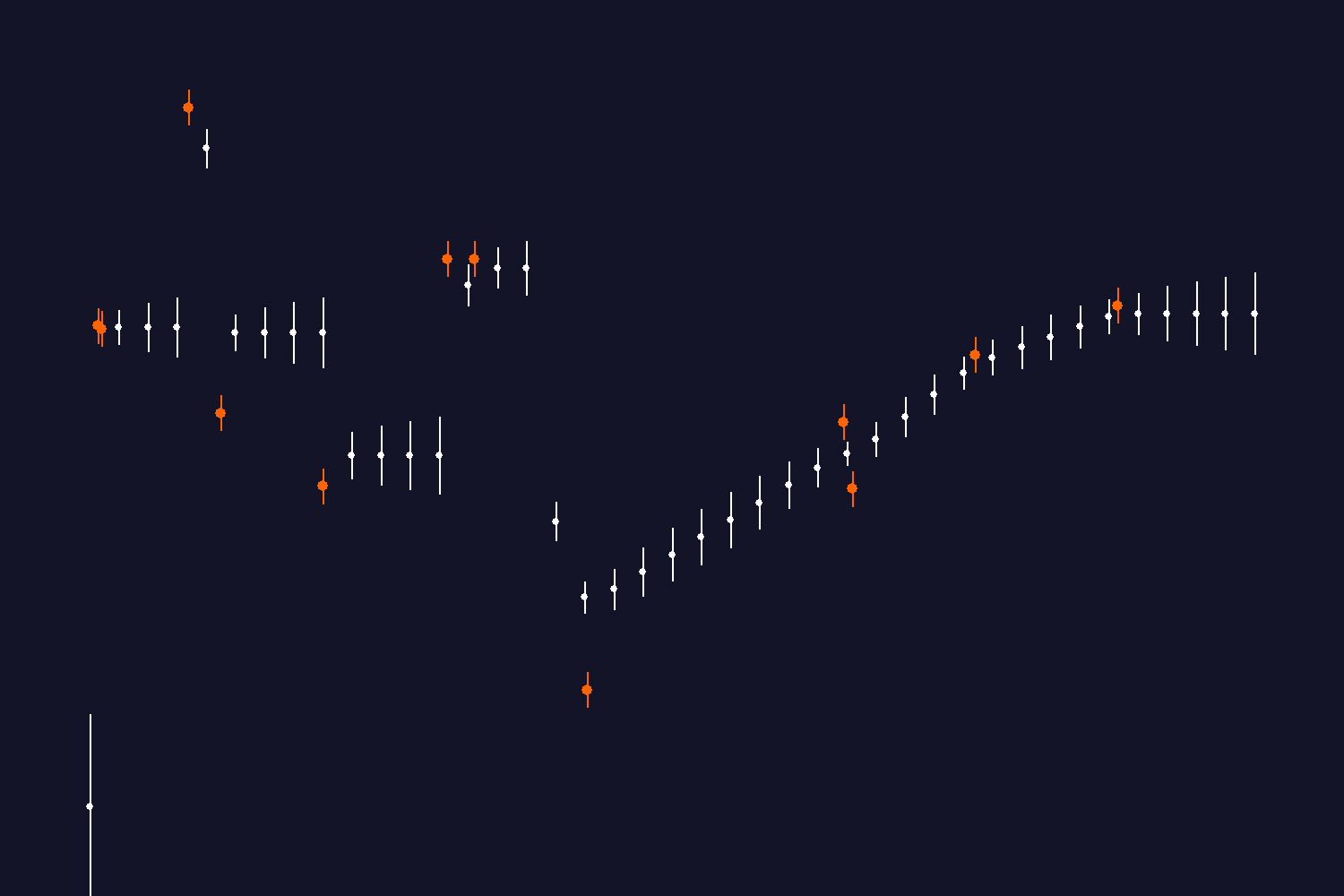}
\hfill
\includegraphics[width=0.245\linewidth]{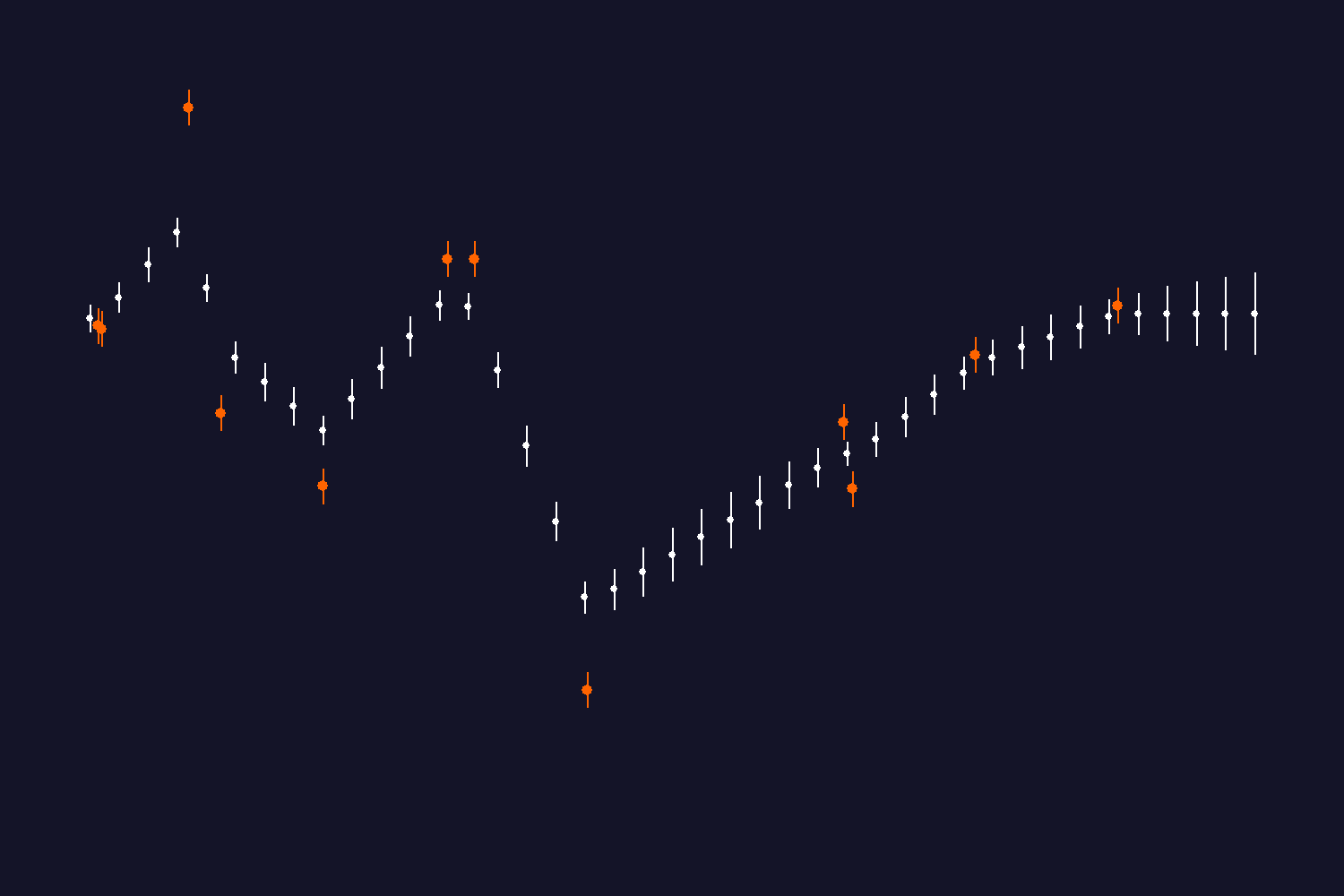}
\hfill
}
\centerline{
\hfill\makebox[0.245\linewidth][c]{\sf Floodfill 48}
\hfill\makebox[0.245\linewidth][c]{\sf Floodfill 80}
\hfill\makebox[0.245\linewidth][c]{\sf Floodfill 130}
\hfill\makebox[0.245\linewidth][c]{\sf Floodfill 160}
\hfill
}
\vspace{1mm}
\centerline{
\hfill
\includegraphics[width=0.245\linewidth]{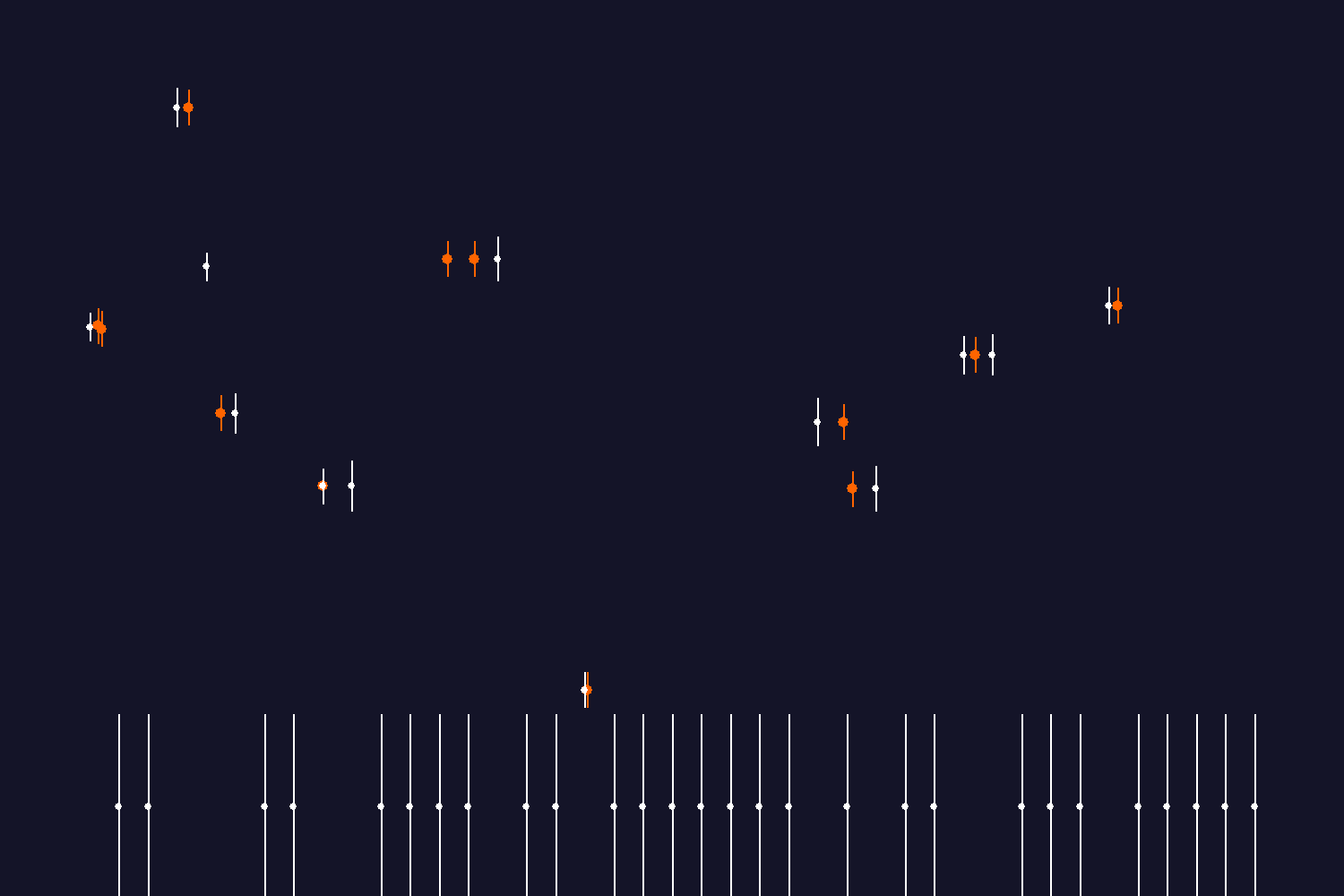}
\hfill
\includegraphics[width=0.245\linewidth]{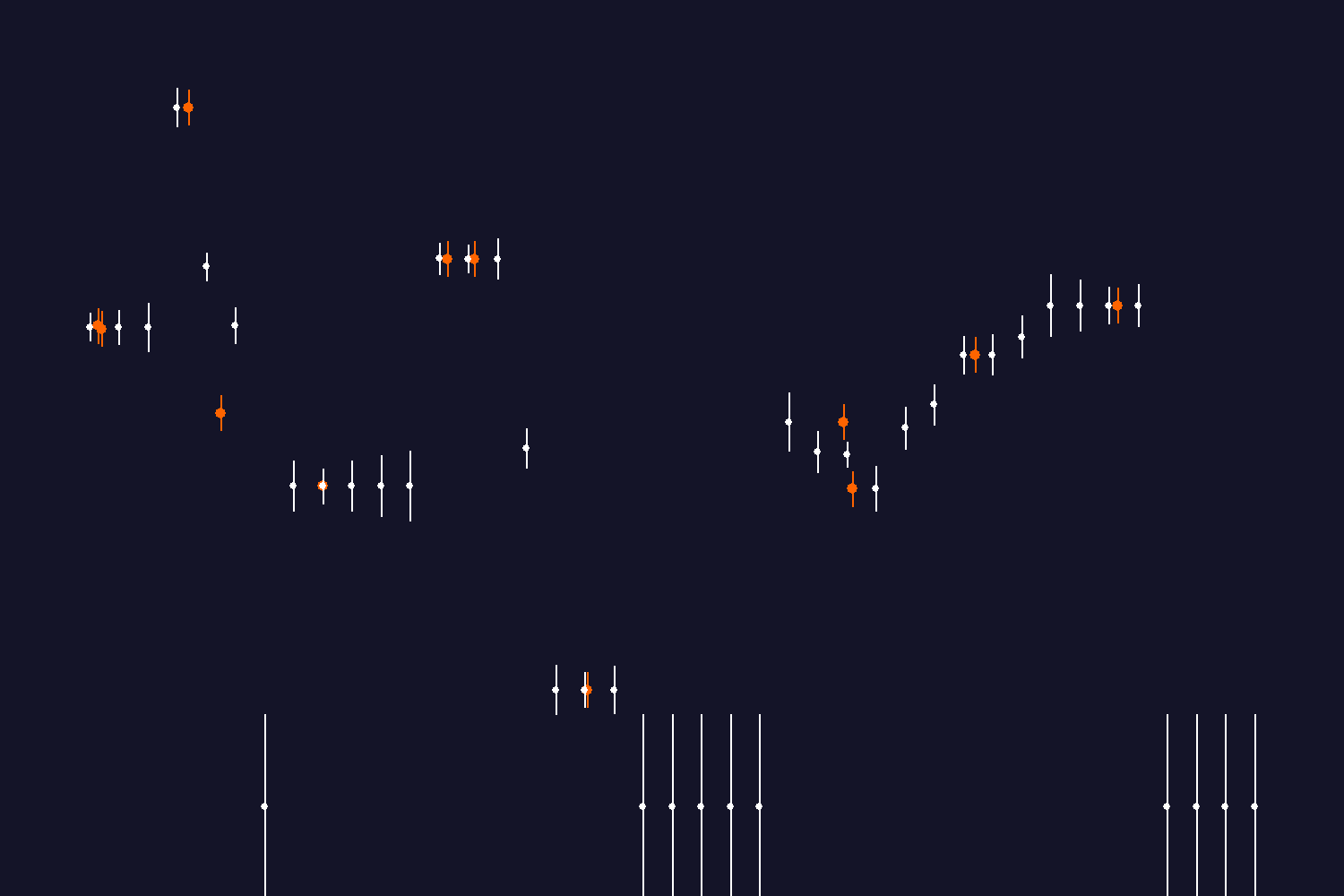}
\hfill
\includegraphics[width=0.245\linewidth]{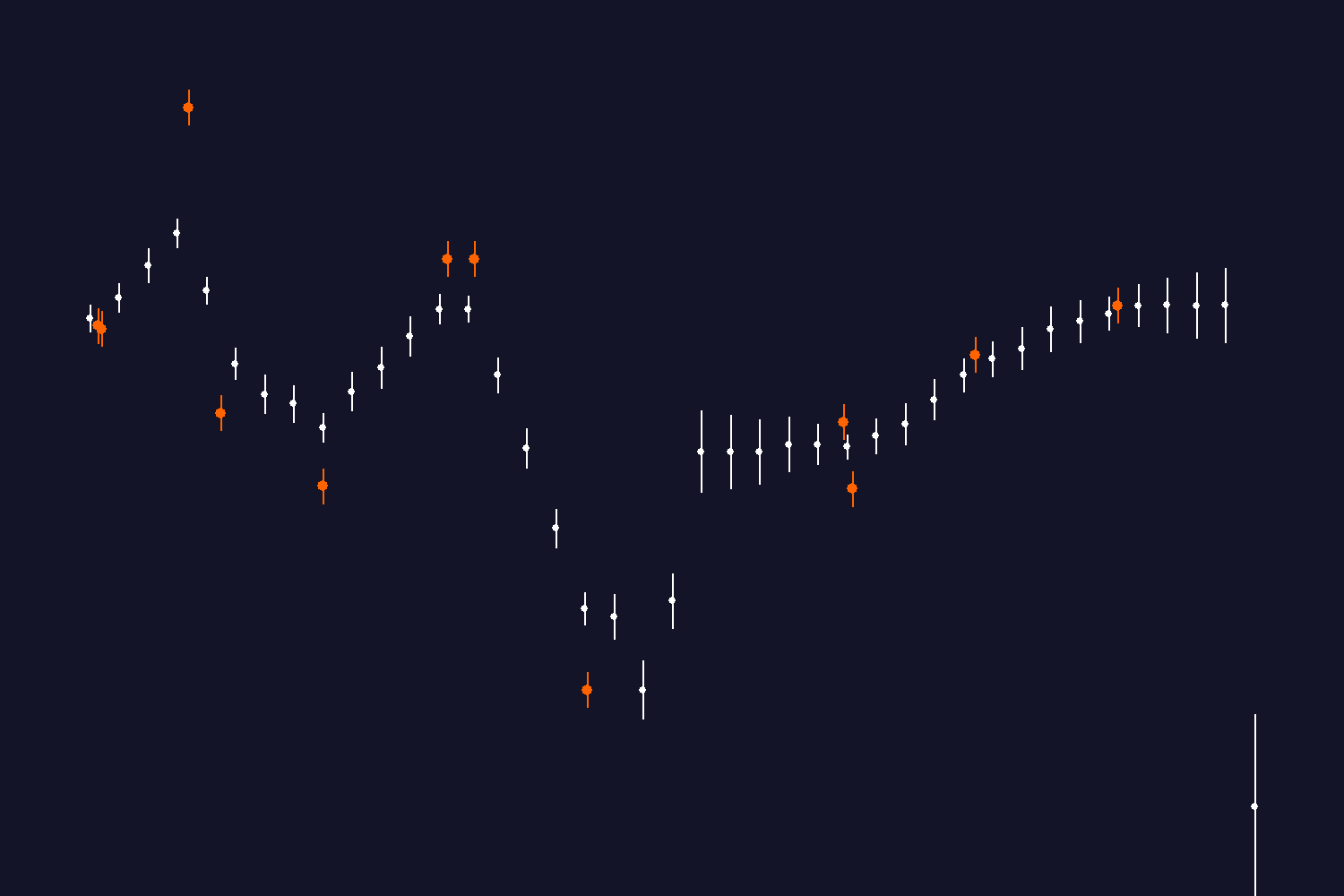}
\hfill
\includegraphics[width=0.245\linewidth]{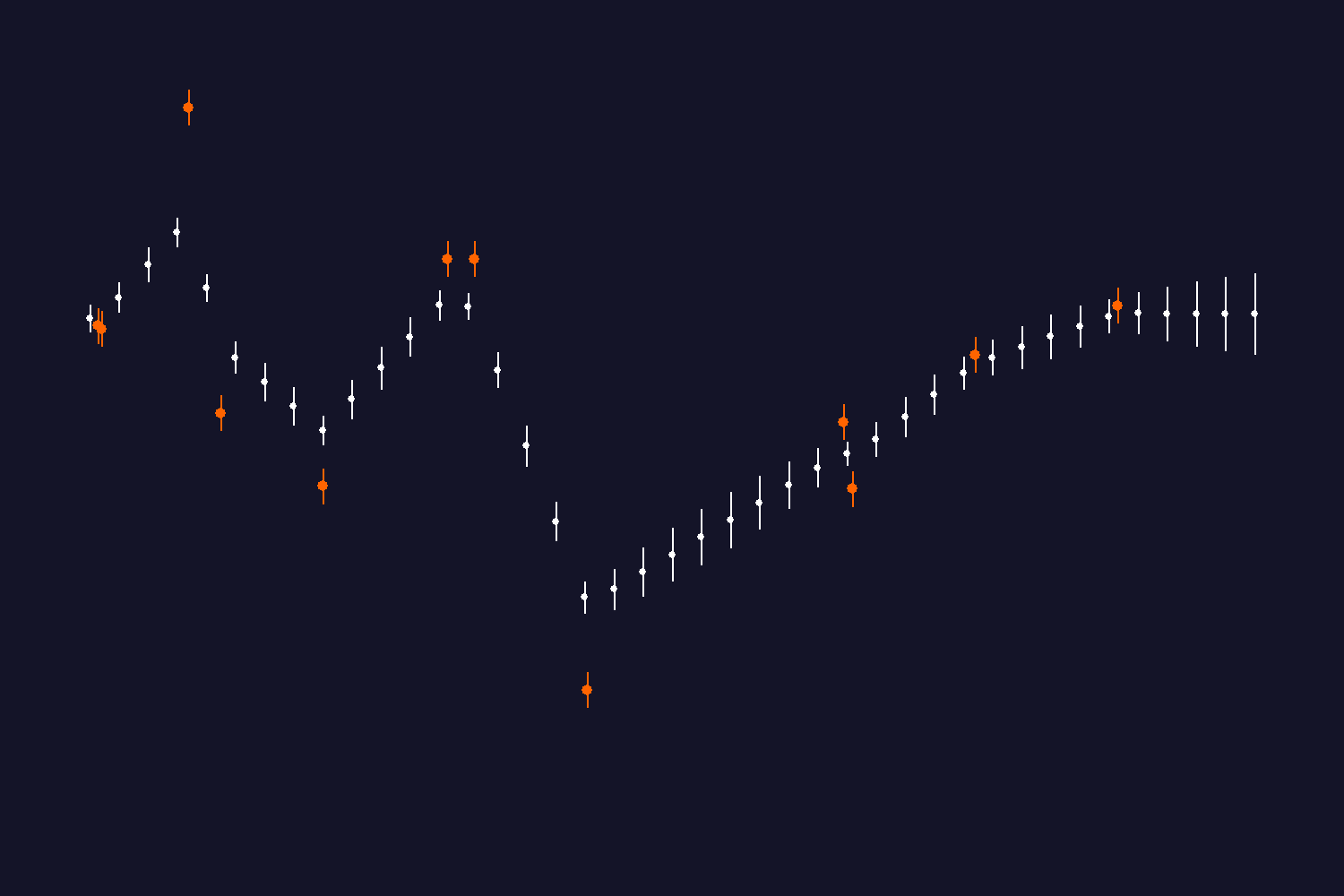}
\hfill
}
\centerline{
\hfill\makebox[0.245\linewidth][c]{\sf Random 243}
\hfill\makebox[0.245\linewidth][c]{\sf Random 670}
\hfill\makebox[0.245\linewidth][c]{\sf Random 2039}
\hfill\makebox[0.245\linewidth][c]{\sf Random 5919}
\hfill
}
\caption{\label{fig:1D}
  1D surface fitting. Measurements are shown with their uncertainty as $\pm 1 \sigma$ error bars in red, and 41 evenly spaced variable state estimates also with $\pm 1 \sigma$ error bars in white.
  Each measurement defines a factor involving the two horizontally closest variables, and a smoothness factor also joins every pair of adjacent variables.
Variables which have not yet received an informative message are initialised to a default zero value with large variance.
  The {\bf top row} shows a serial floodfill message passing schedule, where variables and factors pass messages one by one, first from left to right and then back from right to left. The numbers mean the number of messages that have been passed. After 80, information has propagated all the way from left to right and we see the expected `open ended' variable estimates which are only informed from the left. Messages return back from right to left and by 160 we have reached a full (equivalent to batch) solution, where we see each variable correctly influenced by both measurement and smoothness terms (with larger variances far from measurements).
For comparison, the {\bf bottom row} shows the worst case for message passing of a purely random schedule. At each step we randomly choose one of the 80 edges which connects a variable to a factor, and also randomly choose to pass a message either from factor to variable or from variable to factor. Clearly this is a very inefficient strategy, especially for a serial processor, and many messages must be passed; but eventually after several thousand random messages we reach the same globally correct solution.
}
\vspace{2mm} \hrule
\end{figure*}

In our first example, the goal is to reconstruct a height map surface from a set of point measurements. Each measurement has a perfectly known horizontal position, and Gaussian uncertainty in the vertical direction. We also have a Gaussian smoothness assumption over the surface.

We consider a one-dimensional height map here. We wish to estimate the surface heights at a quantised set of horizontal positions, and define a variable node for each of these. There are an abitrary number of measurements, each of which is associated with a factor node. The smoothness model is a Gaussian constraint on the relative height of every pair of consecutive variables, so there is another set of factor nodes joining these neighbours.
The factor graph of this problem is visualised in Figure~\ref{fig:surfacegraph}.

Each variable node $m_i$ has a fixed $x$ coordinate $x_{m_i}$, and a
height $y_{m_i}$ to be estimated, and therefore a one-dimensional
state variable space:
\beq
\vecx_{m_i} = \vecone{y_{m_i}}
~.
\eeq
Each measurement has a fixed and perfectly known $x$ location
$x_{s}$, and a scalar height measurement $\vecz_s = \vecone{y_s}$. All
height measurements have a fixed measurement precision:
\beq
\matLambda_s =
\matone{\frac{1}{\sigma_m^2}}
~.
\eeq
A measurement at horizontal location $x_s$ is assumed to have a linearly
interpolated dependence on the state variables having $x$ coordinates
$x_{m_1}$ and $x_{m_2}$ which lie either side of it. We define:
\beq
\lambda_s = \frac{x_s - x_{m_1}}{x_{m_2} - x_{m_1}}
\eeq
to be the fraction of the horizontal displacement between the two
variable nodes at which the measurement applies, and therefore deduce
the measurement function:
\beq
\vech_{s}(\vecx_{m_1}, \vecx_{m_2}) = \vecone{(1 - \lambda_s) y_{m_1}
  + \lambda_s y_{m_2}}
~,
\eeq
with Jacobian:
\beq
\matJ_s =
\rowtwo{1 - \lambda_s}{\lambda_s}
~.
\eeq
Additionally, a simple smoothness factor is defined between every pair of consecutive 
variable nodes. We have:
\beq
\vech_{p}(\vecx_{m_1}, \vecx_{m_2}) = \vecone{y_{m_2} - y_{m_1}}
~,
\eeq
with scalar  `measurement' $\vecz_p = \vecone{0}$, Jacobian:
\beq
\matJ_p =
\rowtwo{0}{-1}
~,
\eeq
and fixed precision:
\beq
\matLambda_p =
\matone{\frac{1}{\sigma_p^2}}
~.
\eeq
We linearise the factors, and for each pair of variables we can add
any measurement factors containing them onto the smoothness factor
which already connects them.
This leads to a purely linear graph (variable to factor to variable to factor\ldots) with no loops, and
therefore GBP is known to converge perfectly with one pass in each direction.

We have implemented this example in a simple CPU Python simulator, with a
measurement dataset read in from a text file and interactive
graphics,
available at
\url{http://www.doc.ic.ac.uk/\~ajd/bp1d.py}. The code requires a
straightforward Python3 installation with NumPy for numerics and PyGame for
interactive visualisation.

Figure~\ref{fig:1D} shows screenshots from the simulator, and the
caption explains the progress of GBP estimation for two types of
message passing schedule which represent the two extremes of
efficiency for serial processing. In the top row of figures, we see a
`floodfill' scenario, where messages start from the leftmost variable
and are passed one by one all the way along the chain of variables and factors to
the far right, then all the way back. With one full traversal in both
directions, all variables and factors are fully `informed' from all
parts of the graph, and we achieve the globally optimal solution.
In the bottom row, we see the progress instead of a fully random
message passing schedule, where many steps will incur
wasted work if the variable or factor sending the message has not
itself updated since its last message. However, the most interesting
thing to observe here is that full convergence to the global optimum
is still reliably reached after some thousand iterations, with purely random
and distributable processing.

\subsection{2D Pose Graph}
\label{section:constraint}

\begin{figure}[t]
	\centerline{
		\hfill
		\includegraphics[width=\columnwidth]{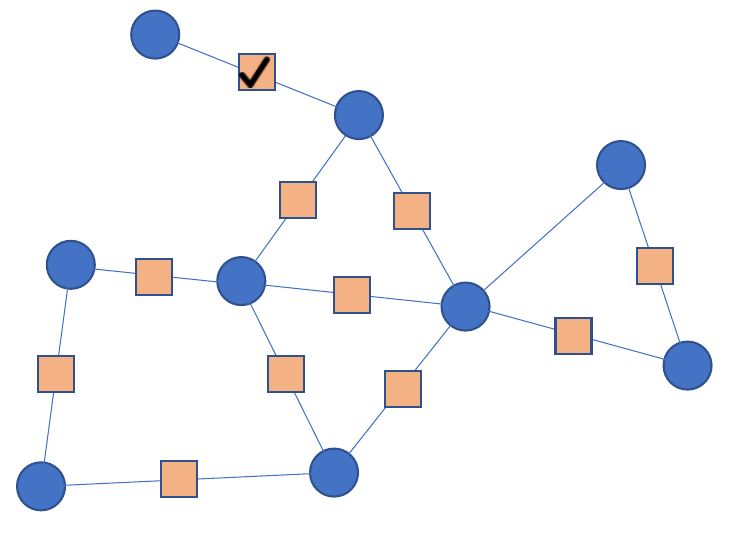}
		\hfill
	}
	\caption{\label{fig:2Dgraph}
		Factor graph for a 2D pose graph problem. Variable nodes (blue) scattered around the 2D plane are linked by randomly  generated measurement factors (orange) between nearby nodes. Note that in our implementation we also use some  absolute pose unary factors on the variables (one strong, to anchor one variable, the others very weak), which are not visualised here. 
	}
	\vspace{2mm} \hrule
\end{figure}

Our second example is a
linear 2D pose graph, a simple version of the pose graph optimisation problem in mobile robotics where many robots or a single exploring robot must estimate their global locations from a network of purely relative measurements. 
This application of GBP is still linear, but now loopy, and therefore
iterative message passing will certainly be needed to approach a
global solution.

The factor graph under consideration is shown in Figure~\ref{fig:2Dgraph}.
In detail, each variable node has two degrees of freedom for its position on the
2D plane:
\beq
\vecx_{m_i} = \vectwo{x_{m_i}}{y_{m_i}}
~.
\eeq
Each factor is a 2D Euclidean relative pose measurement $\vecz_s$ between two
nodes, with measurement function:
\beq
\vech_{s}(\vecx_{m_i}, \vecx_{m_j}) = \vecx_{m_j} - \vecx_{m_i}
~,
\eeq
and fixed measurement precision:
\beq
\matLambda_s =
\mattwo{\frac{1}{\sigma_m^2}}{0}{0}{\frac{1}{\sigma_m^2}}
~.
\eeq
Measurement function $\vech_{s}$ is linear so we only need to
construct the linear constraint once. Given the Jacobian:
\beq
\matJ_s = \mattwofour{-1}{0}{1}{0}{0}{-1}{0}{1}
~,
\eeq
we construct information vector $\veceta_s$ and precision matrix
$\matLambda_s'$ using Equation~\ref{eqn:lin}. Note that when the
measurement function is linear as it is here, $\matJ_s \vecx_0 =
\vech_s(\vecx_0)$ and those two terms cancel out, so simply $\veceta_s = \matJ_s^\top \matLambda_s
 \vecz_s$.

 We set up a simulation (code for our Python simulator is available
 from \url{http://www.doc.ic.ac.uk/\~ajd/bpmap.py}). We define 20 variable nodes, which have randomly
 generated ground truth locations on the 2D plane. 50 measurements are
 also generated, each of which randomly connects two variable nodes,
 and which has a randomly generated measurement value sampled from a
 Gaussian
 distribution around the ground-truth value with precision $\matLambda_s$.
  We also add a weak unary pose factor to each variable node. These
 factors have identity measurement functions and Jacobians.
One chosen variable has a much stronger unary pose
factor. This means that the whole network will be anchored.

\begin{figure*}[t]
\centerline{
\hfill
\includegraphics[width=0.33\linewidth]{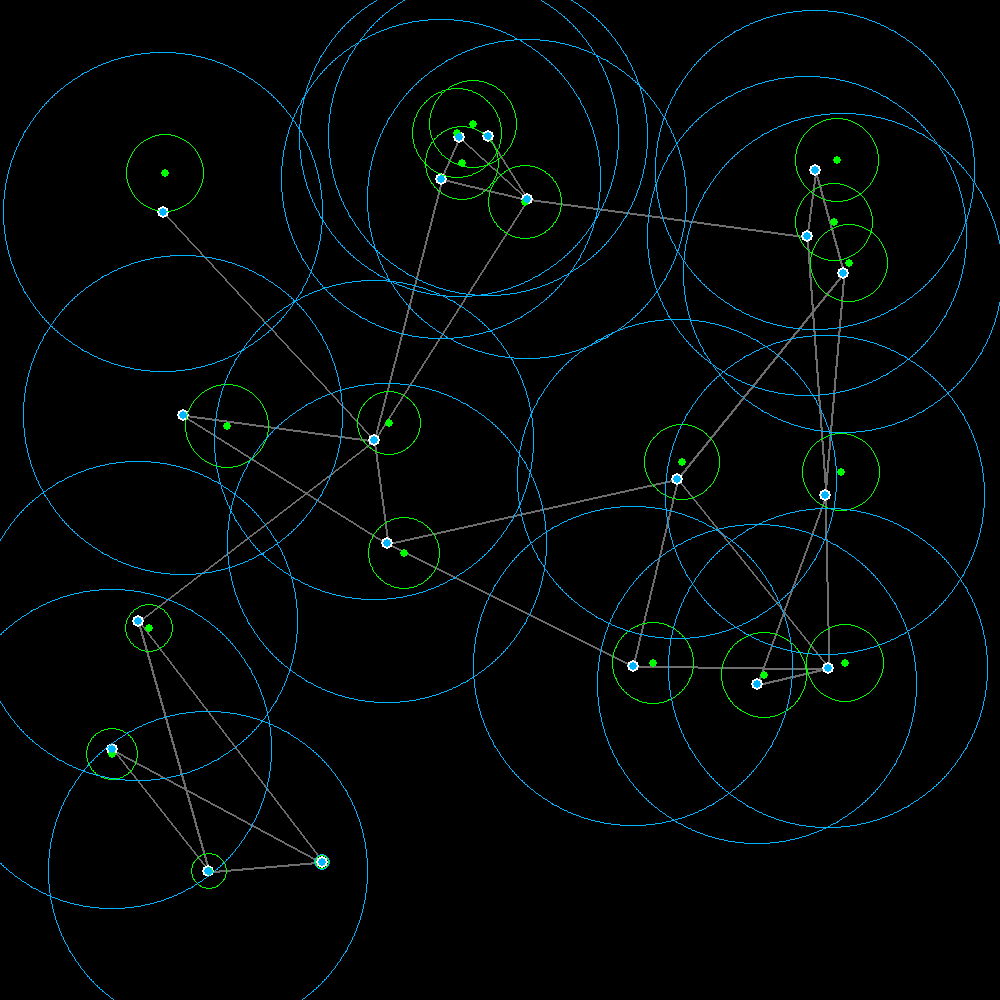}
\hfill
\includegraphics[width=0.33\linewidth]{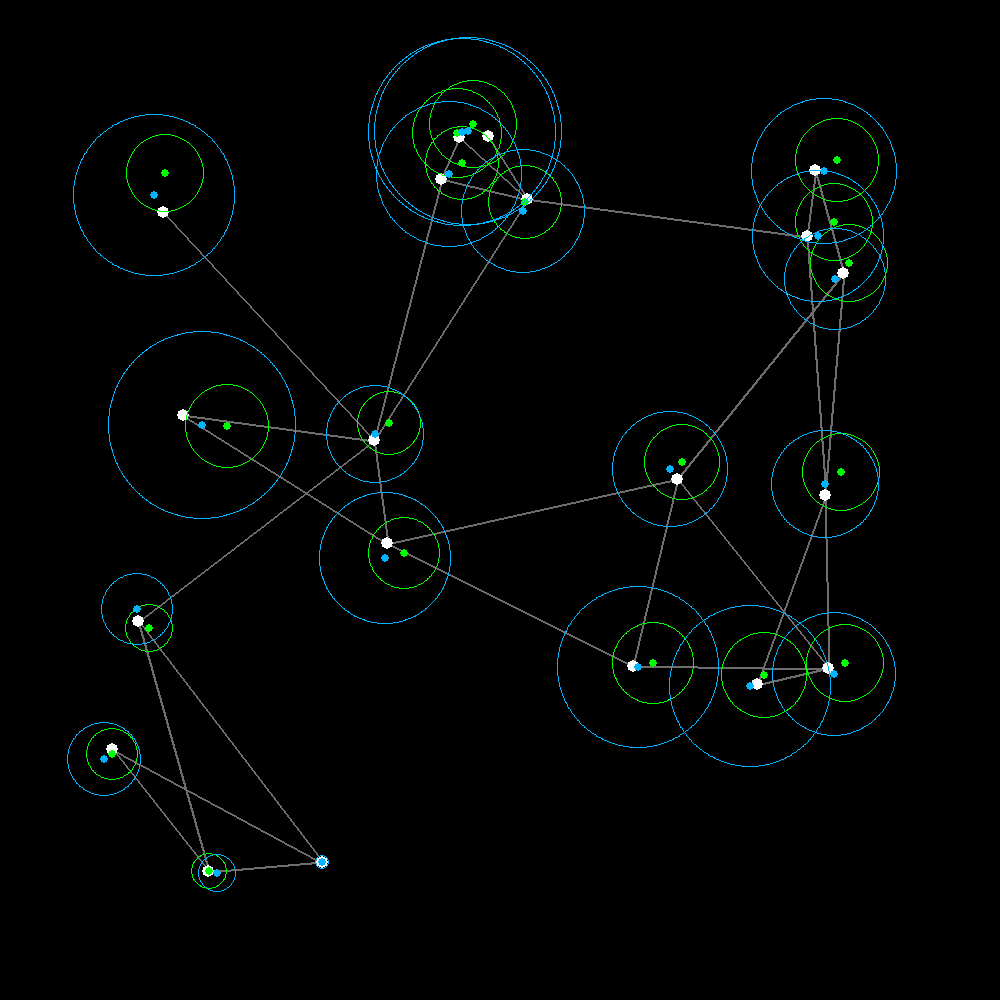}
\hfill
\includegraphics[width=0.33\linewidth]{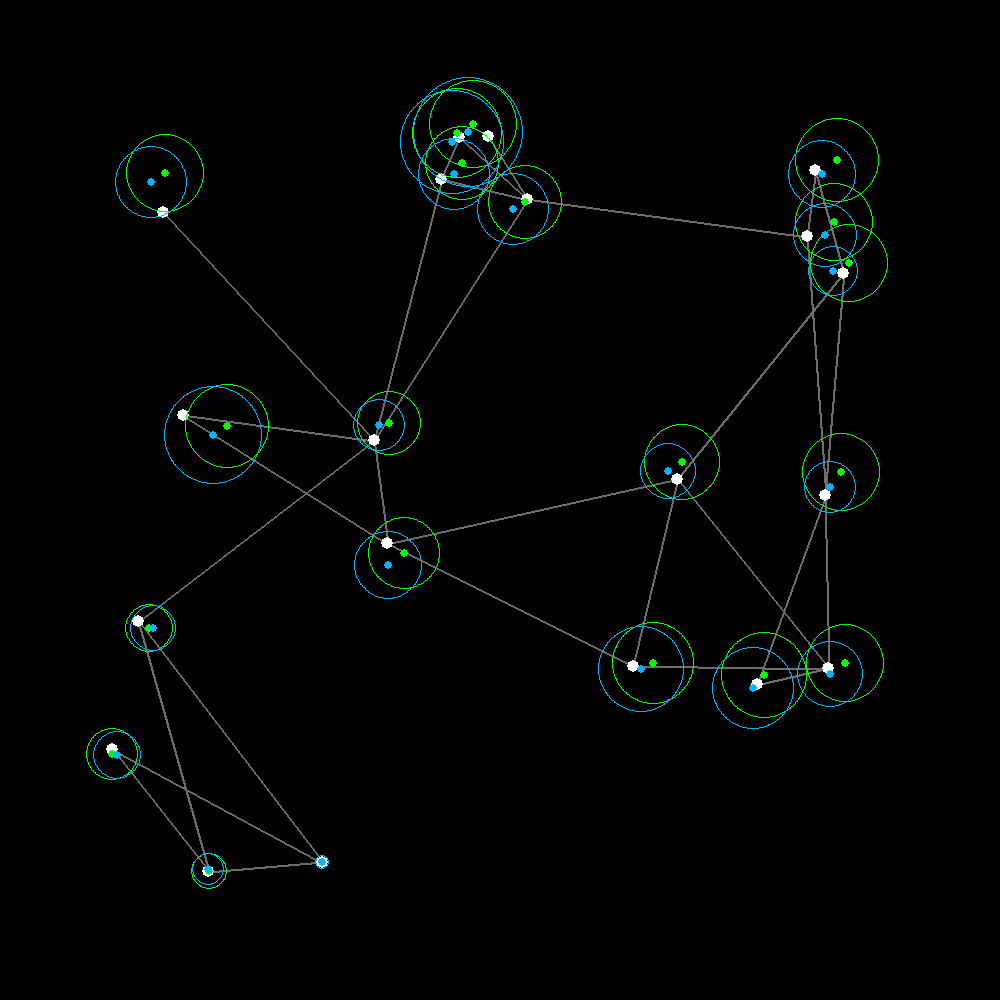}
\hfill
}
\centerline{
\hfill\makebox[0.33\linewidth][c]{\sf 0 steps}
\hfill\makebox[0.33\linewidth][c]{\sf 1 step}
\hfill\makebox[0.33\linewidth][c]{\sf 2 steps}
\hfill
}
\vspace{1mm}
\centerline{
\hfill
\includegraphics[width=0.33\linewidth]{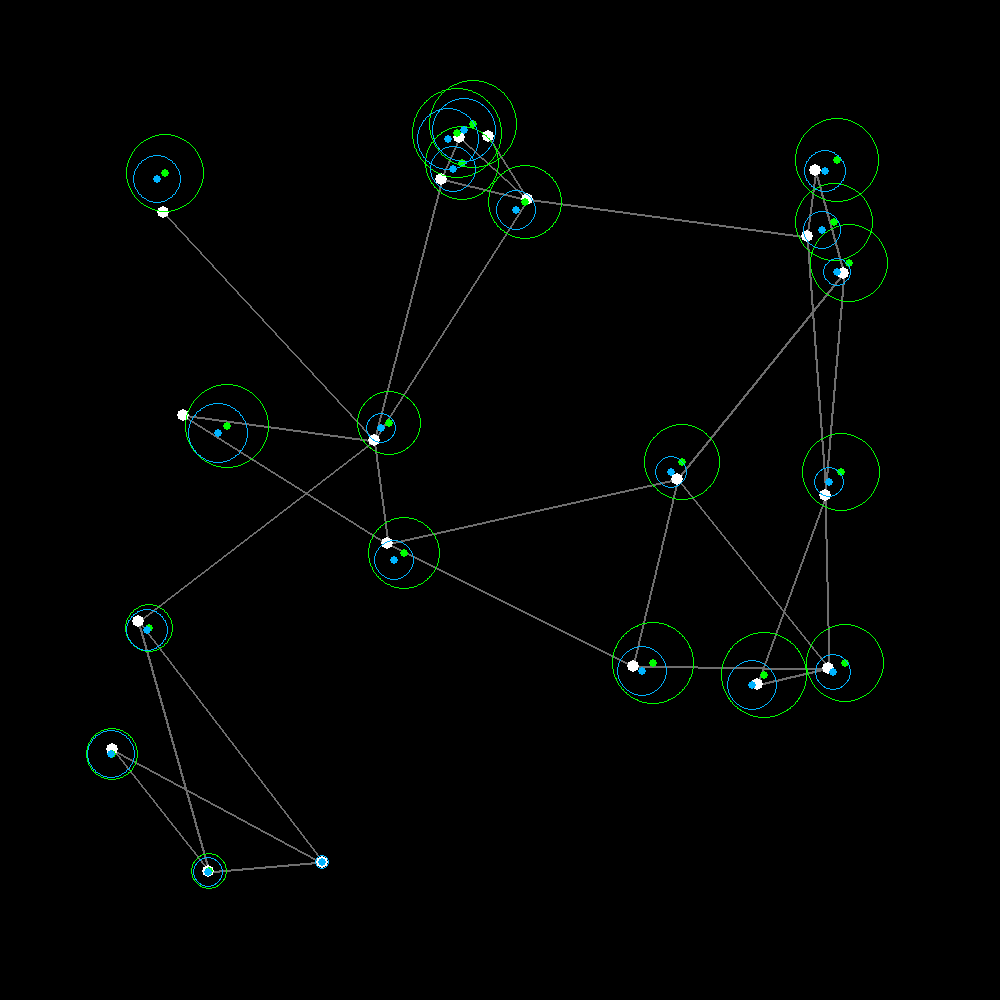}
\hfill
\includegraphics[width=0.33\linewidth]{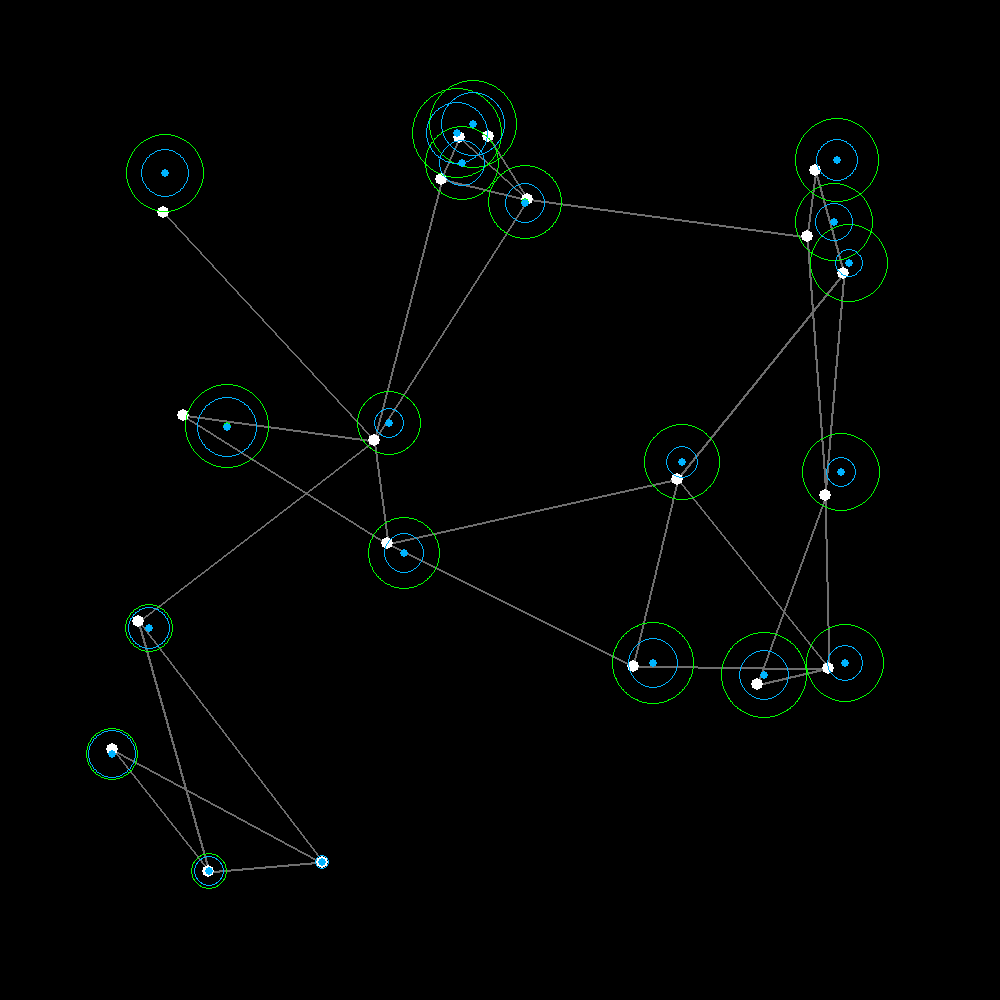}
\hfill
\includegraphics[width=0.33\linewidth]{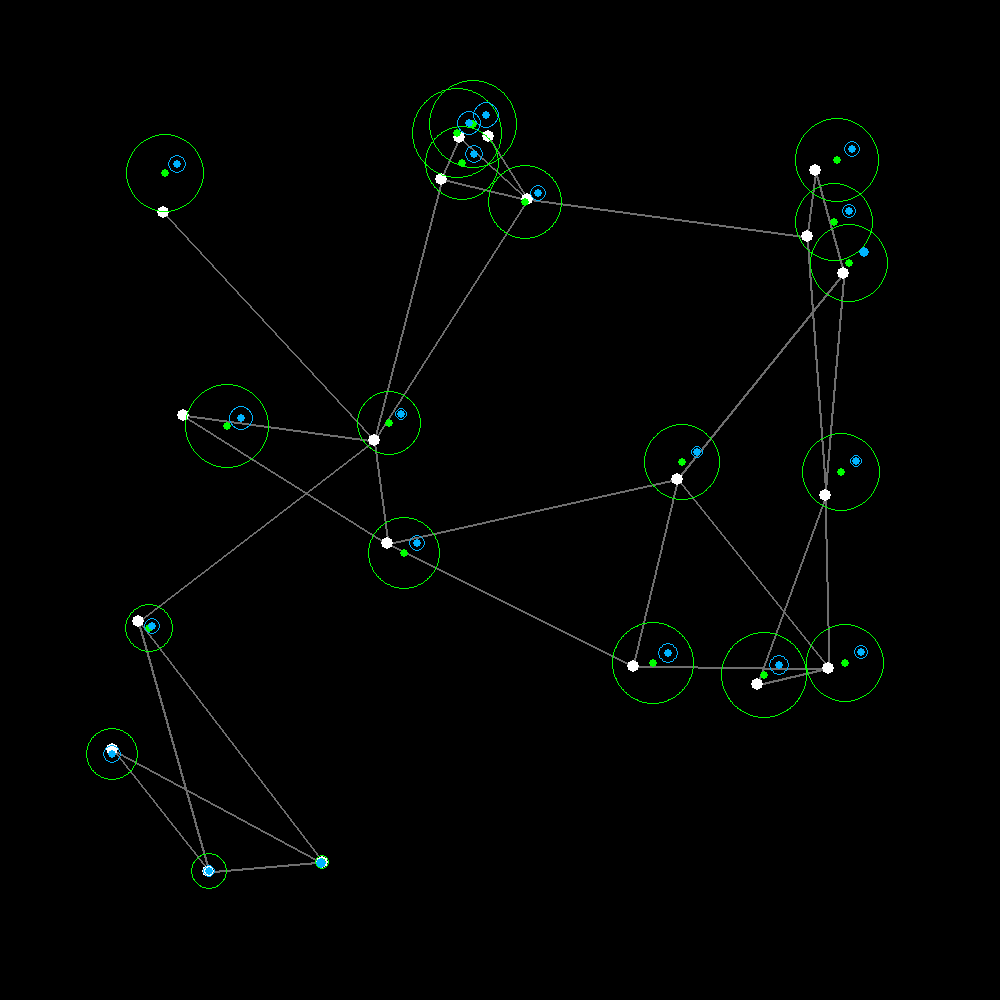}
\hfill
}
\centerline{
\hfill\makebox[0.33\linewidth][c]{\sf 6 steps}
\hfill\makebox[0.33\linewidth][c]{\sf 171 steps}
\hfill\makebox[0.33\linewidth][c]{\sf Measurement Precision Increased}
\hfill
}
\caption{\label{fig:2dtest}
2D linear pose graph, showing iterations of a random message
passing scheme. The ground truth locations of variable nodes are shown
in white, with estimates in blue (mean point locations and one sigma
covariance as a circle). Each grey line is a measurement factor
connecting two variables. There are 20 variable nodes and 50
measurements. Each variable also has a weak absolute pose factor,
except for the anchored variable at the bottom centre of the scene which has a
strong absolute pose factor, and these pose factors pass messages
once  to their variables at initialisation. Message passing then
proceeds in the steps shown, simulating a parallel message schedule
where at each step first every variable passes a message to all of the
measurement factors it connects to, and then every measurement factor
sends a message to every variable it connects to.
In each sub-figure, on top of the incremental estimation we also show
in green the batch solution to the full pose graph problem (mean
point and one single covariance as a circle). We see that only a few
steps are needed for the graph to reach good relative estimates
(e.g. after 6 steps, where we see an almost constant shift between the
GBP and batch solutions for nearby nodes).
After 171
steps, information has fully propagated from the anchored node and the mean estimates from GBP are indistinguishable from the
batch solution, though the covariances are in many cases overconfident.
In the final panel we show that after convergence (or at any other
time) we can make dynamic and local changes to the factors, the
effects of which are transmitted to the whole graph with no changes to
the GBP algorithm. Here we increase the precision of all measurement
factors which changes the mean estimates because the weak pose factors are
trusted less. (Note that we did not account for this in the green
batch solution shown, which does not change.) 
}
\vspace{2mm} \hrule
\end{figure*}

So each variable node has zero or more randomly connected measurement
factors. We visualise the progress of a simulated parallel message
passing schedule in Figure~\ref{fig:2dtest}, and discuss its progress
in the caption, including a comparison with a full batch
solution. This batch solution is formed by adding all linearised
factors into a single large information matrix, and inverting this
matrix to find the mean and covariance of all variables.
 
We observe that in this small graph, it takes a rather large number of iterations for the means to converge
to estimates which are indistinguishable from the batch solution, but
we should note that relative estimates in the graph are obtained after
fairly few steps. For instance, if we  look at the result after 6
steps in Figure~\ref{fig:2dtest}, nodes nearby in the graph have mean
estimates which differ from the batch solution by very similar offsets.
At the stage the estimates in the graph are most likely highly useful
for any application where relative information is important, such as
robot navigation. The difference between 6 steps and 171 steps is that
the rooted node with a strong pose factor is finally able to propagate absolute information
around the whole graph.
The covariances
of the estimates are overconfident, which is a well known property of
GBP, with generally greater overconfidence for nodes which are highly
inter-connected graph regions.

In the final panel of the figure, we give an example of the extreme flexibility of GBP estimation, and the ability of decentralised estimation methods to work in an editable, reversible way. After convergence, we change the precision of all of the factor nodes to a stronger value, and the result of this quickly propagates through the graph, with no global coordination needed. Any number of dynamic changes like this can easily be dealt with, which we  think will be important  in the future of Spatial AI, where for instance some human input or machine learning process produces an updated value of a prior assumption (e.g. surface smoothness) which with most estimation methods would have become `baked into' the representation.

\subsection{Related Work}

Pose graph optimisation is a heavily studied problem in robotics and
computer vision, and the state of the art in academia is represented
by excellent open source libraries such as 
g2o
\cite{Kummerle:etal:ICRA2011},
GTSAM
\cite{Dellaert:Kaess:Foundations2017}
and
Ceres
\cite{CeresManual}. These libraries run on the CPU, and use methods
such as sparse Cholesky decomposition to efficiently factorise the
invert the information matrix of large problems. We make no claims
that GBP is competitive with such methods on a CPU, because it
requires many iterations. As noted before, the structure of doing
processing with a CPU and unified memory storage allows `god view'
analysis of an optimisation problem and the determination of very
efficient solvers. Presumably industry has in-house versions of these
which are even more efficient.

Our argument in favour of GBP is about its suitability for different
computing architectures such as graph processors with fully
distributed processing and storage; and also its high flexibility and
usefulness in handling the heterogeneous, always-changing graphs in
Spatial AI.
On hardware like a graph processor,  our measure of what is an
efficient algorithm needs to change from the CPU standard of total
computation time. The performance of an algorithm on a graph processor
depends on its need for distributed computation, storage and data
tranmission, and probably a suitable measure should be multi-dimensional.

Within well known methods for CPU pose graph optimisation, the ones that
are closest to GBP with a random message schedule are those based on 
Stochastic Gradient Descent, such as by 
Olson~\etal~\cite{Olson:etal:ICRA2006} or 
Grisetti~\etal~\cite{Grisetti:etal:ITITS2009}. We have not had the
chance to make a head to head performance comparison with these first order
methods, though GBP certainly has some appealing positive points even
on the CPU, such as no need for the tuning of gain constants.

\section{Robust Factors using M-Estimators}

Most practical estimation methods in computer vision and robotics take
account of the fact that sensors, especially outward-facing ones like
cameras, have a measurement probability distribution which is not
truly Gaussian. The classic behaviour is that the distribution is
closely Gaussian when the sensor is essentially `working', and
reporting measurements that are close to ground truth apart from small
variations due to quantisation and similar, but then some percentage
of the time the sensor will report wildly incorrect `garbage'
measurements. For instance, if a camera is reporting the image
location of matched image features, false correspondences happen
sometimes and give measurements arbitrarily far away from ground truth.
If we plot the measurement distribution of such a sensor we see a
distribution which looks like a Gaussian centrally but is more `heavy-tailed'.

In optimisation and estimation, such behaviour is modelled using a
family of `robust' functions called M-Estimators. Here we will show
that these robust functions can be easily incorporated into GBP with
completely local processing.

Consider a standard measurement factor defined as in
Equation~\ref{equ:generalfactor} and linearised as in
Equation~\ref{eqn:lin}. 
The form is:
\beq
f_s(\vecx_s) = K e^{-\frac{1}{2} E_s}
~,
\eeq
where $E_s$, the least squares `energy' of the constraint, is:
\beq
E_s = (\vecz_s - \vech_s(\vecx_s))^\top \matLambda_s (\vecz_s -
\vech_s(\vecx_s))
~.
\eeq

The term:
\beq
\label{eqn:mahalonobis}
M_s = \sqrt{
(\vecz_s - \vech_s(\vecx_s))^\top \matLambda_s (\vecz_s -
  \vech_s(\vecx_s))
}
\eeq
is the Mahalonobis distance, representing the number of standard
deviations that the measurement is away from the mean of the
distribution, and so for a standard Gaussian constraint we use a
simple square $E_s =
M_s^2$.
In robust estimation, we vary this by setting a threshold level on
$M_s$ beyond which we change the energy to a function which rises less
steeply.

Let us first consider the commonly used Huber function which transitions
from quadratic to a straight line beyond a threshold $M_s \geq N_\sigma$.
A factor with Huber loss has the following form:
\beq
f_s(\vecx_s) =
\begin{cases}
K e^{-\frac{1}{2} M_s^2} &\ M_s \leq N_\sigma \\
K e^{-(N_\sigma M_s  - \frac{1}{2} N_\sigma^s)} &\ M_s \geq N_\sigma
\end{cases}
~,
\eeq
such that the two parts of the function match up in terms of both
value and gradient at the discontinuity $M_s = N_\sigma$.

Now, in GBP, every message takes the form of an information vector
and precision matrix representing a Gaussian distribution. So what we
do to represent the effect of the non-Gaussian part of a robust factor
is to find the Gaussian distribution which has the same
value energy, and pass a message with that precision instead.
This is similar to the Dynamic Covariance Scaling method in~\cite{Agarwal:etal:ICRA2013}.
We ask what Mahalonobis distance we must be from the mean in a
standard quadratic energy to be equivalent to the Huber energy in the
linear region. Specifically, we need to find $M_{sR}$ such that:
\beq
\frac{1}{2} M_{sR}^2 = N_\sigma M_s  - \frac{1}{2} N_\sigma^s
~.
\eeq
Rearranging we find:
\beq
M_{sR} = \sqrt{2 N_\sigma M_s - N_\sigma^2}
~.
\eeq
And therefore:
\beq
\label{eqn:kR}
k_R = \frac{M_{sR}^2}{M_s^2} = \frac{2 N_\sigma}{M_s} - \frac{N_\sigma^2}{M_s^2}
\eeq
is the factor by which the energy of the constraint should be
reduced. Remembering the information form of
Equation~\ref{eqn:coninf}, we see that this is achieved by multiplying
both the precision matrix $\matLambda_s'$ and information vector
$\veceta_s$ by this factor.

So, to summarise, to use a Huber norm on a factor, every time that
factor is to pass a message we first use {\em all} the latest incoming
messages from variables in order to form its state vector $\vecx_s$,
and then evaluate the current Mahalonobis distance $M_s$ using
Equation~\ref{eqn:mahalonobis}. We test this against the $N_\sigma$
cutoff we have set for this factor (which might be 4.0 or something
similar), representing the number of standard deviations from the mean
for which we expect Gaussian behaviour. If $M_s \leq N_\sigma$ we
are in the Gaussian zone and use the standard linearised precision
matrix and information vector for the message calculate. If $M_s \geq
N_\sigma$,
we temporarily scale 
$\matLambda_s'$ and 
$\veceta_s$ by factor $k_R$ as calculated in Equation~\ref{eqn:kR} for
the purposes of this message pass only.

We can use the same method to handle other robust norms. For instance,
a function which is Gaussian up to $M_s \leq N_\sigma$ and then constant
beyond is implemented with a factor $k_R = \frac{N_\sigma^2}{M_s^2}$.

We will see that these robust factors allow lazy data association
during GBP (reminiscent of \cite{Olson:Agarwal:IJRR2013}), where the robust status of factors can change dynamically
during ongoing graph optimisation, giving the ability to reject poor
measurements immediately or after enough contradictory alternative
data has been received.

An interesting future area for research is a multi-modal approach
where we might initialise multiple robust factors with different
precision values to represent a single measurement, and allow them to
`fight it out' over iterations of BP to find the best supported
hypothesis, and achieving a discrete model-selection capability.

\section{Incremental SLAM}

\begin{figure}[t]
	\centerline{
		\hfill
		\includegraphics[width=\columnwidth]{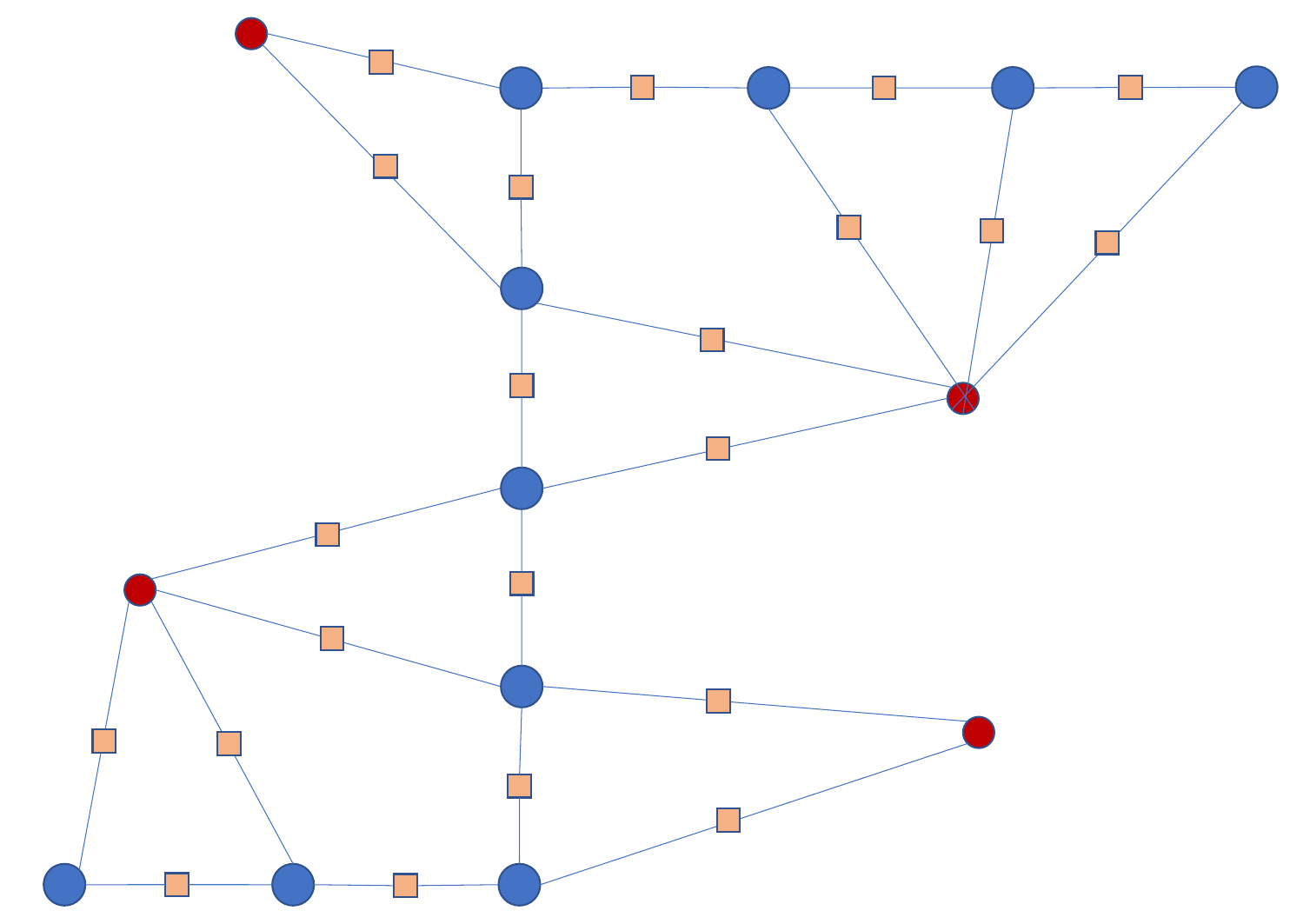}
		\hfill
	}
	\caption{\label{fig:robotgraph}
		Factor graph for a 2D SLAM problem. Blue nodes are variable nodes describing the robot position while red nodes are landmark variable nodes. The orange squares are factors representing 2D constraints. 
	}
	\vspace{2mm} \hrule
\end{figure}

\begin{figure*}[t]
\centerline{
\hfill
\includegraphics[width=0.49\linewidth]{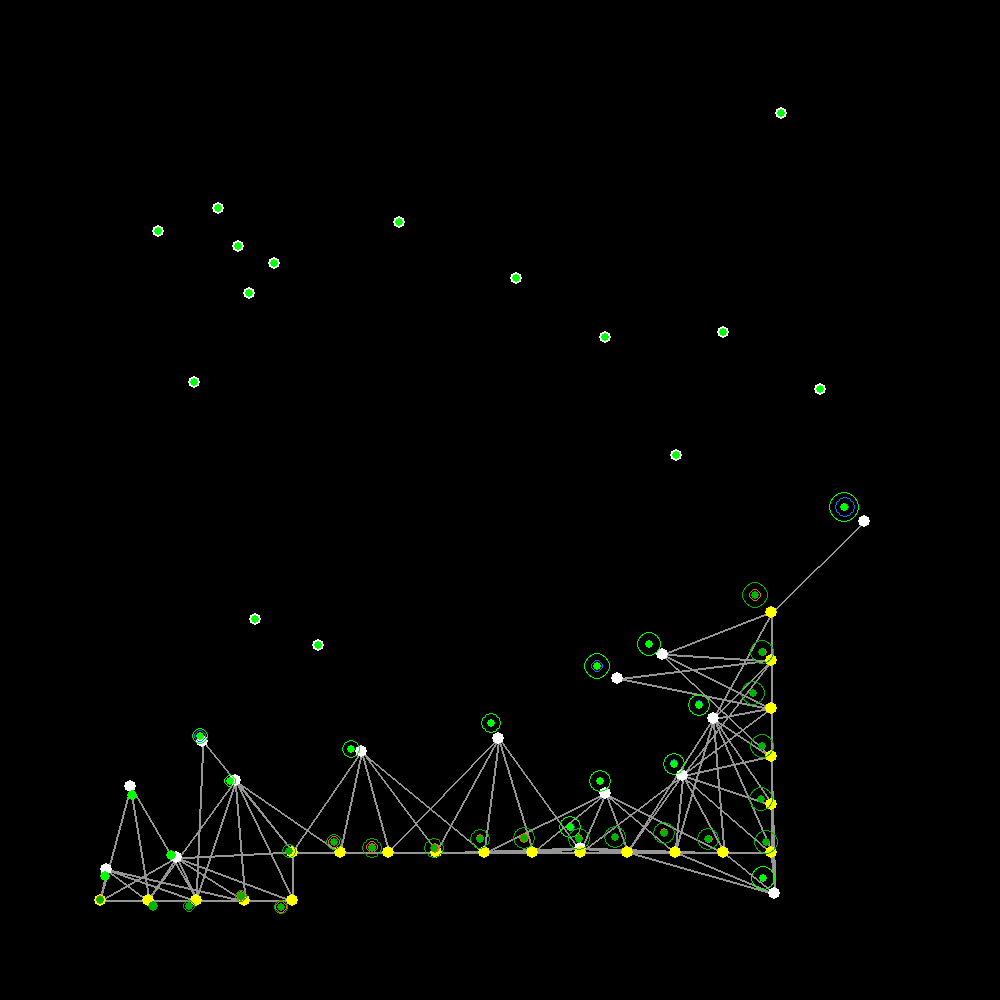}
\hfill
\includegraphics[width=0.49\linewidth]{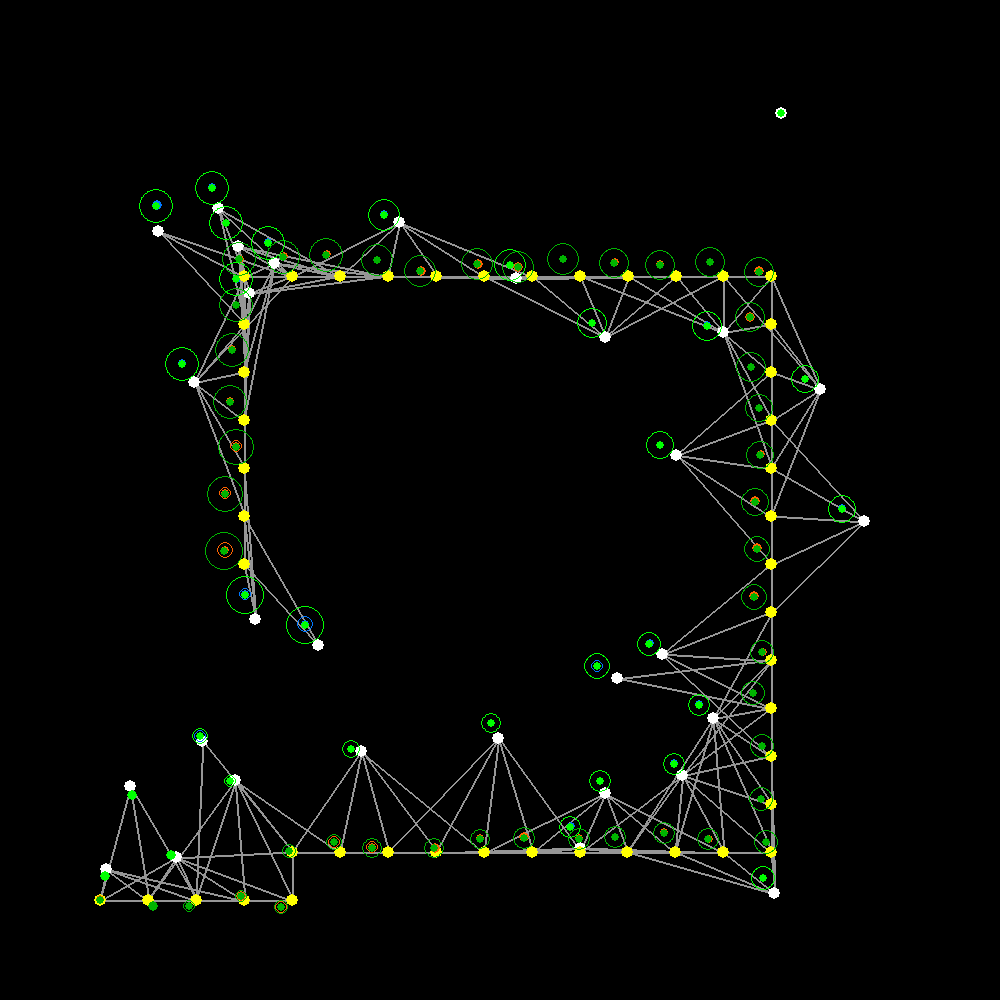}
\hfill
}
\centerline{
\hfill\makebox[0.49\linewidth][c]{\sf Early exploration}
\hfill\makebox[0.49\linewidth][c]{\sf Just before loop closure}
\hfill
}
\vspace{1mm}
\centerline{
\hfill
\includegraphics[width=0.49\linewidth]{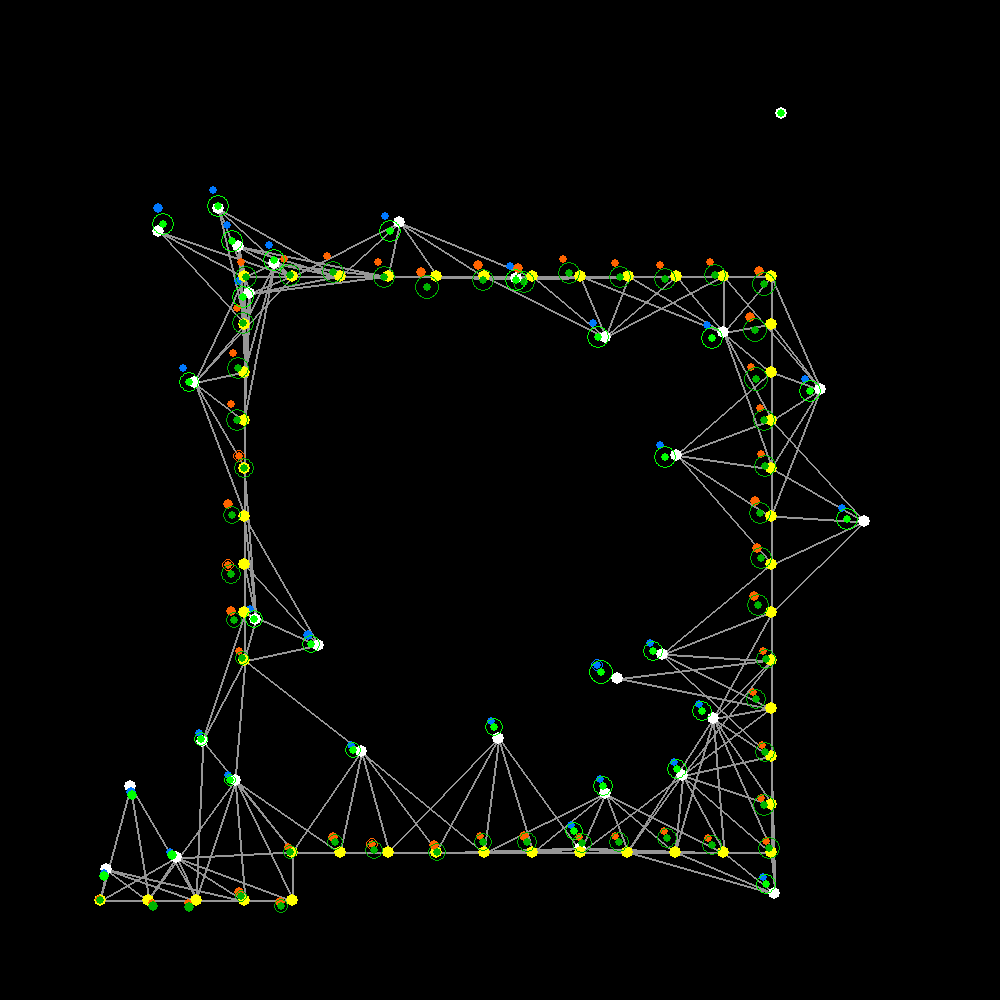}
\hfill
\includegraphics[width=0.49\linewidth]{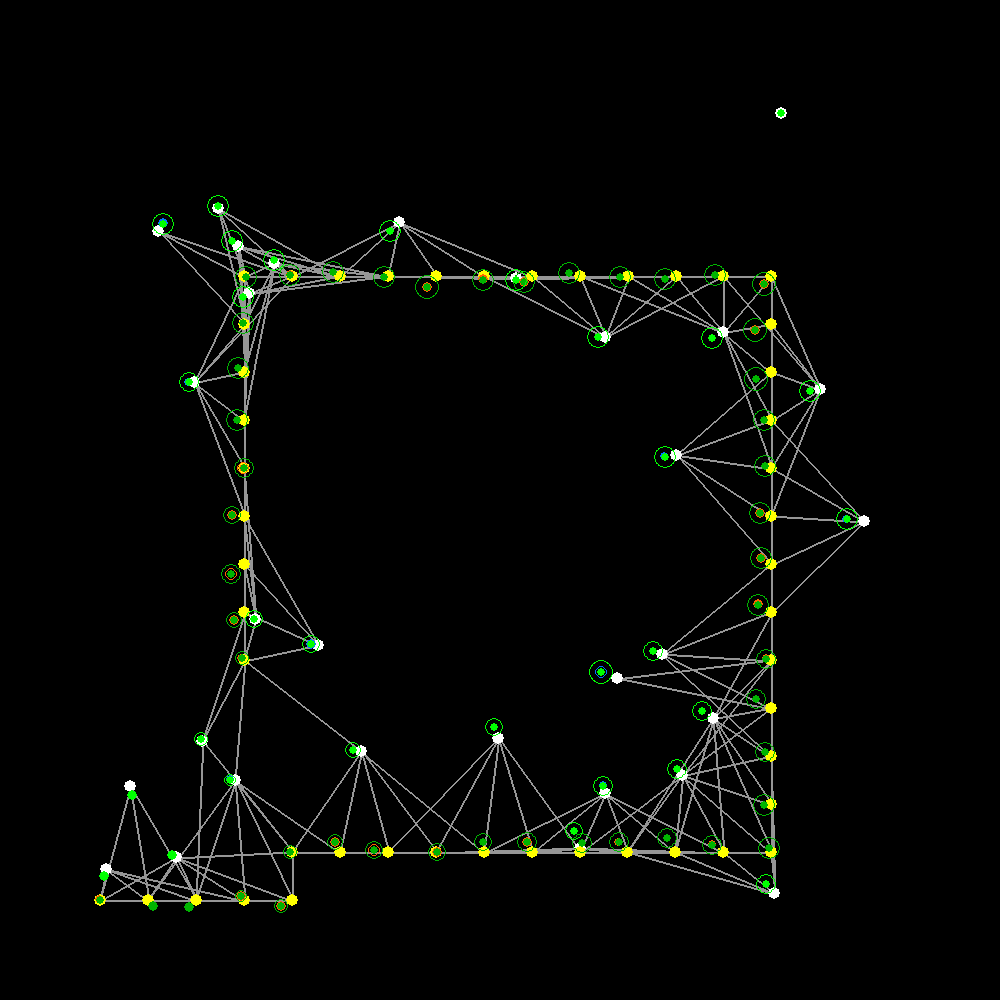}
\hfill
}
\centerline{
\hfill\makebox[0.49\linewidth][c]{\sf Just after loop closure}
\hfill\makebox[0.49\linewidth][c]{\sf Steady state convergence}
\hfill
}
\caption{\label{fig:2dslam}
2D SLAM simulation. A moving `robot', with history of ground truth
poses shown as yellow dots, explores a scene with containing
landmarks (ground truth positions are white dots), and from each
pose makes observations of nearby landmarks to incrementally build the
factor graph shown in grey. Red and blue dots and ellipses show the
robot and landmark estimates obtained from GBP, which runs
continually on the growing factor graph. We also superimpose the
optimal batch solution to the graph in green for comparison. We see
close agreement between the GBP and batch estimates until loop
closure occurs, where the large correction needed takes a large number
of GBP iterations to propagate fully around the graph. However, even
very soon after loop closure we see that the relative information in
the GBP estimate is good, with nearby nodes having estimates
differing from the batch solution by similar amounts.
}
\vspace{2mm} \hrule
\end{figure*}

We will now show how 
GBP can be straightforwardly applied to an ever-changing SLAM graph,
including the optional use of robust factors to account for poor data
association.

\subsection{Incremental SLAM with Standard Gaussian Factors}

First we will tackle incremental SLAM but using standard Gaussian factors.
We simulated a 2D cartesian SLAM problem, where as a `robot'
translates it leaves a history of pose variable nodes, with each
consecutive pair joined by a factor on their relative locations
representing a measurement from odometry (see Figure~\ref{fig:robotgraph}). Scattered throughout the
simulated 2D environment are landmarks the robot can observe. From
each new robot pose, factors are added to the graph to represent
measurements of the landmarks withing a bounded distance. All
measurements in the simulation have randomly sampled Gaussian noise,
using different but constant covariances for the odometry factors and
measurement factors respectively; the factors in the graph are
initialised with the corresponding precision matrices.
There is no rotation in the simulation, and all measurements are in
cartesian space, so this is again a formulation where the dependence
of measurements on variables is purely linear, and the mathematical
details of variable and factor message passing are the same as in the
2D constraint graphs of Section~\ref{section:constraint}.
We have a strong pose factor attached to the first robot variable
node, anchoring this node and effectively defining the coordinate
frame for SLAM.

We visualise the progress of SLAM estimation in
Figure~\ref{fig:2dslam}. In our simulation, which is available from
\url{http://www.doc.ic.ac.uk/\~ajd/bpslam.py}, keyboard controls w,a,s,d can be used to move the robot, and the
factor graph is generated automatically and incrementally. In the
background, we run a continuous schedule of simulated parallel message passing.
At each step, all variables use their waiting incoming messages to
calculate outgoing messages to all connected factors. Then, in
alternation, all factors do the same thing. The variables and factors
in the dynamically changing graph are stored in dynamic vector data
structures and we can easily iterate through all of them as the vectors grow.
The figure shows that we make SLAM estimates which are consistent with a batch solution, and that GBP can comfortably cope with the dynamically changing graph, including major events such as loop closure.

\subsection{Incremental SLAM with Robust Factors}

\begin{figure*}[t]
\centerline{
\hfill
\includegraphics[width=0.33\linewidth]{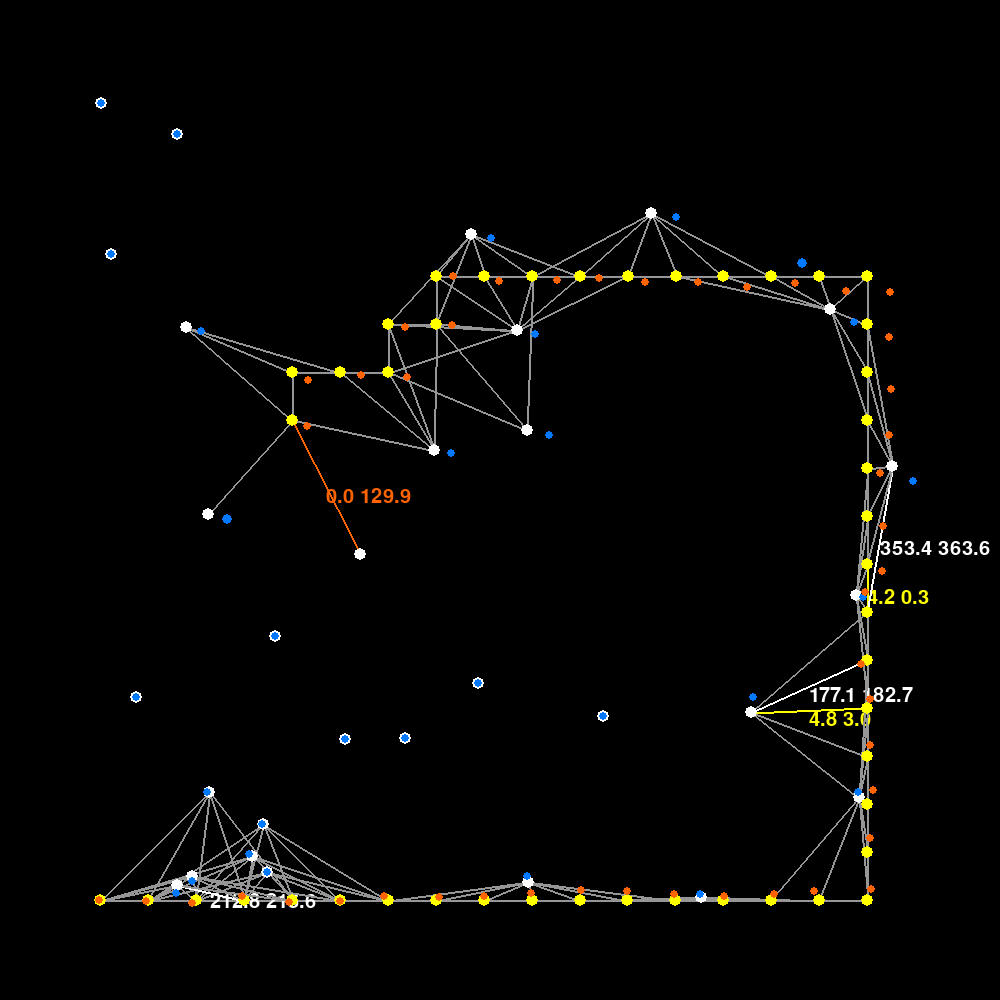}
\hfill
\includegraphics[width=0.33\linewidth]{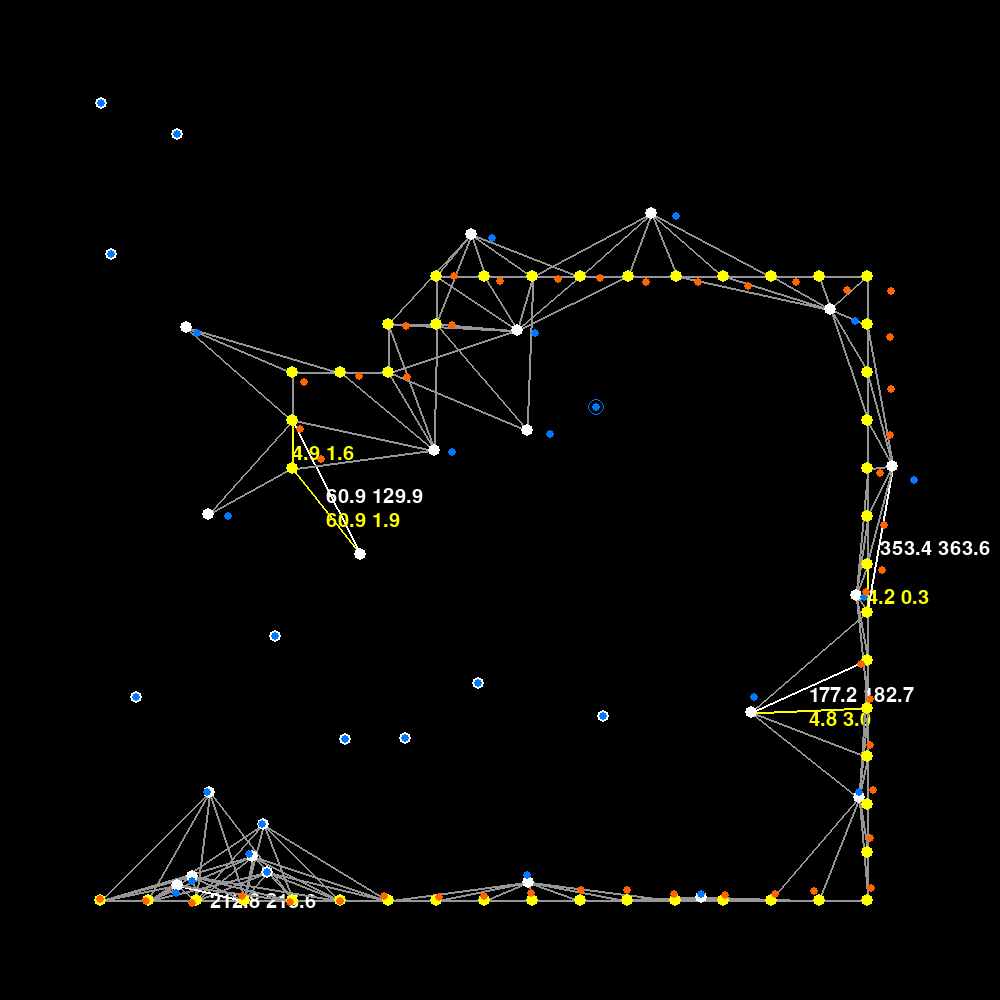}
\hfill
\includegraphics[width=0.33\linewidth]{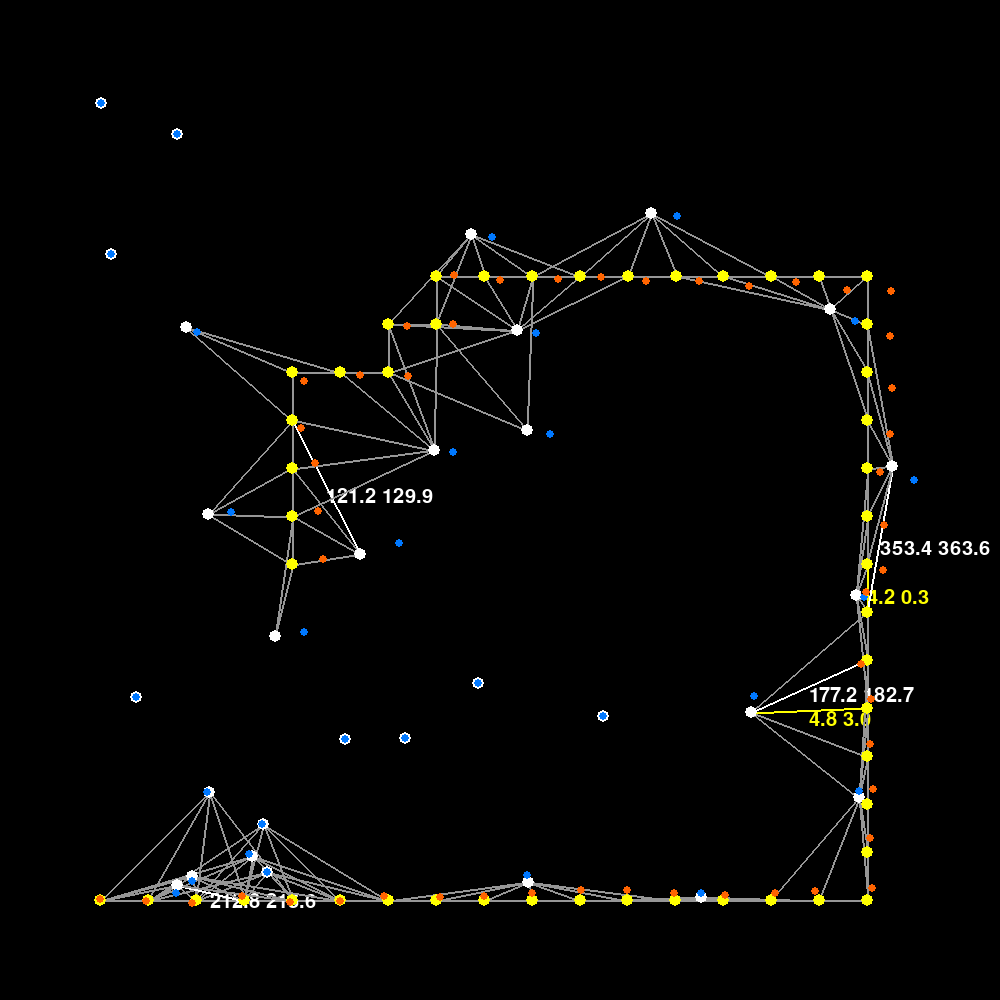}
\hfill
}
\centerline{
\hfill\makebox[0.33\linewidth][l]{\small \sf New erroneous measurement (red)}
\hfill\makebox[0.33\linewidth][l]{\small\sf Second measurement of landmark}
\hfill\makebox[0.33\linewidth][l]{\small\sf Further good measurements; error rejected}
\hfill
}
\centerline{
\hfill\makebox[0.33\linewidth][l]{\small \sf}
\hfill\makebox[0.33\linewidth][l]{\small\sf White and yellow factors balance error}
\hfill\makebox[0.33\linewidth][l]{\small\sf Erroneous factor is white; others grey}
\hfill
}
\vspace{1mm}
\centerline{
\hfill
\centerline{
\hfill
\includegraphics[width=0.33\linewidth]{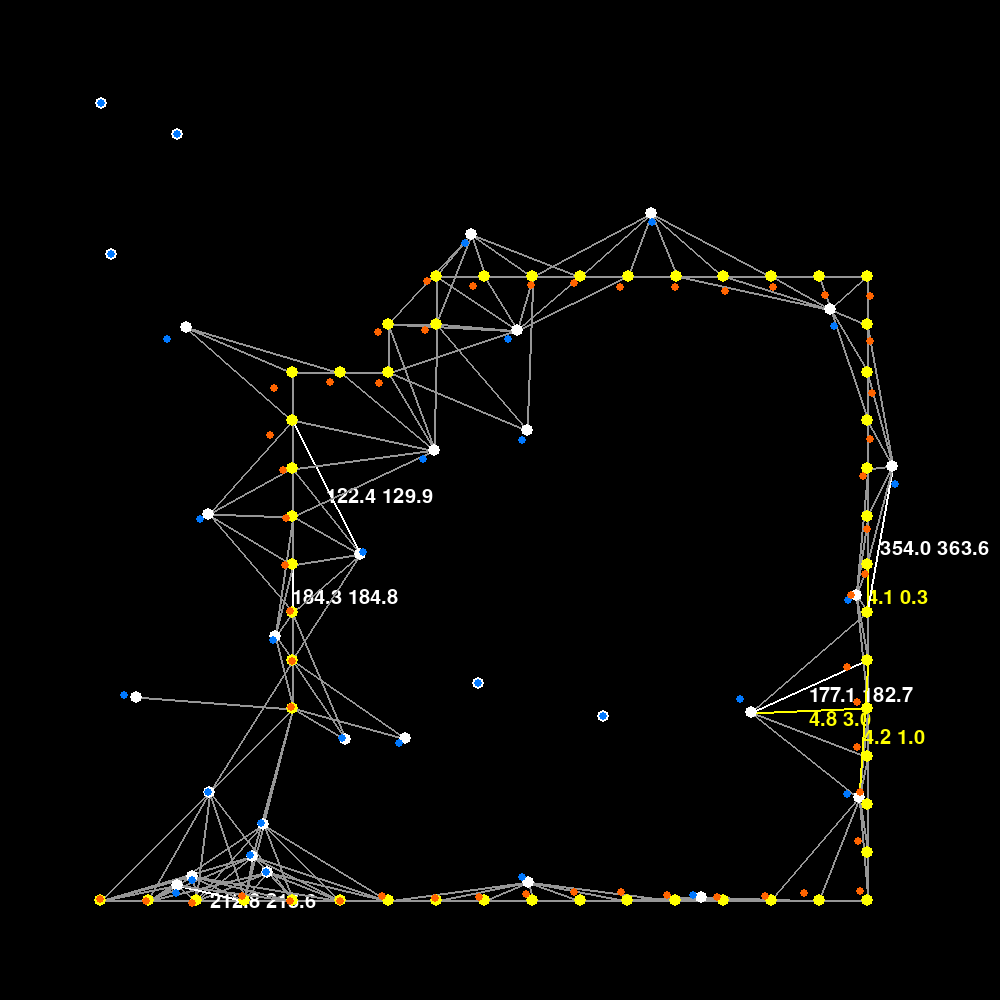}
\hfill
\includegraphics[width=0.33\linewidth]{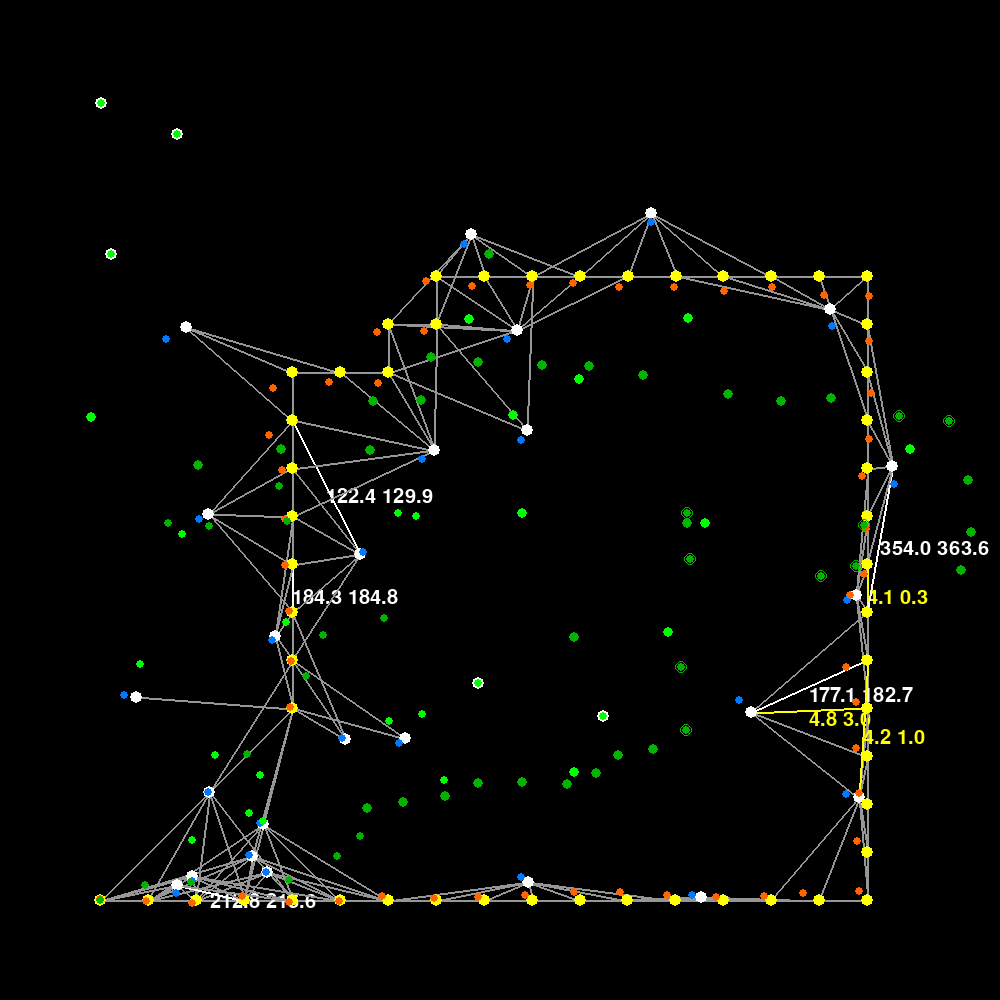}
\hfill
\includegraphics[width=0.33\linewidth]{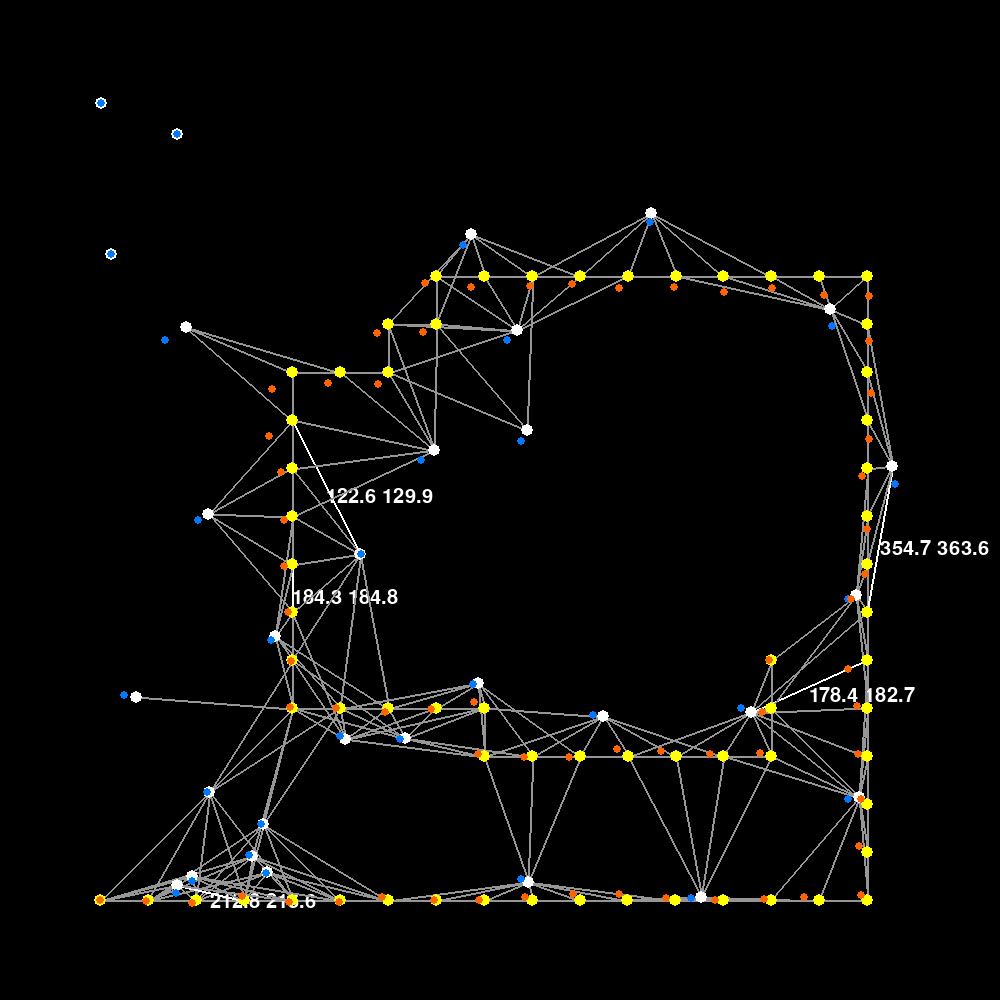}
\hfill
}
}
\centerline{
\hfill\makebox[0.33\linewidth][l]{\small\sf Loop closure}
\hfill\makebox[0.33\linewidth][l]{\small\sf Comparison with non-robust }
\hfill\makebox[0.33\linewidth][l]{\small\sf After more measurements, }
\hfill
}
\centerline{
\hfill\makebox[0.33\linewidth][l]{\small\sf}
\hfill\makebox[0.33\linewidth][l]{\small\sf batch solution (green)}
\hfill\makebox[0.33\linewidth][l]{\small\sf 4 outliers confidently identified}
\hfill
}
\caption{\label{fig:2drobust}
  2D SLAM simulation with random erroneous measurements and robust
  factors. Every time a factor passes a message, it reevaluates its
  Mahalonobis distance. In this simulation, we use a Huber loss with
  Mahalonobis distance threshold of 4.0 to transition from quadratic
  to linear cost. One in 50 ground truth measurements in the
  simulation has a large error added. In the visualisation, factors
  are colour-coded: {\bf grey} if the Mahalonobis distances of both
  the ground truth measurement and the current factor estimate are
  both below 4.0 (normal case); {\bf white} if both distances are
  above 4.0 (an erroneous measurement that BP has recognised as such
  and is treating with the linear part of the Huber function; {\bf
    red} for a factor with high ground truth distance but low in the
  factor: this is an erroneous measure not yet recognised; and {\bf
    yellow} for a factor with high factor distance but low  ground
  truth distance; this is a good measurement being `unfairly' treated
  as an outlier. All non-grey factors display their estimated and true
  Mahalonobis distances numerically. When a first erroneous
  measurement is made to initialise a new landmark, there are no
  contradictory good measurements to oppose it. In the second and
  third panels, additional measurements of the landmark support each
  other and push the erroneous measurement far into the linear Huber
  region. Good estimates are seen throughout the graph after loop
  closure, and a much better result than the green batch solution
  which does not take account of outliers (of course, we could have
  used robust methods there as well). More exploration and further
  loop closure happens by the last panel, and four erroneous
  measurements are now very confidently identified.
}
\vspace{2mm} \hrule
\end{figure*}

In Figure~\ref{fig:2drobust} we now see the performance of GBP for
incremental SLAM when a random $\frac{1}{50}$ of all measurements have a large
error added, and all factors now use a robust Huber kernel. This can
be tried out as part of the same simulation
\url{http://www.doc.ic.ac.uk/\~ajd/bpslam.py}, pressing `r' to
enter robust mode. GBP with robust factors has the impressive
capability to detect outlier measurements in a local and lazy manner,
with erroneous measurements which were not immediately apparent often
determined much later when enough support builds up for a better
hypothesis. Playing with the simulator is the best way to get a good
feel for this.

\subsection{Towards Front-End Use}

Our example here of SLAM with robust factors shows how errors that
slip through the front-end measurement part of a SLAM system (such as
visual feature matching) could be cleaned up by back-end
estimation. In the longer term, we are interested in how GBP and
graph-based estimation could be used for the whole of a Spatial AI
system, including front-end data association like the matching in a
sparse or dense SLAM system, avoiding the need for ad-hoc algorithms
such as RANSAC. We believe that this will be possible, and plan to
experiment further soon. For instance, in dense SLAM data associate
(such as the ICP tracking in
KinectFusion~\cite{Newcombe:etal:ISMAR2011}), each pixel measurement
from a depth camera is associated with several possible locations in a
dense model, and this association is refined iteratively through
ICP. We could replace this process with GBP, where the measurement might be
connected by several factors to different scene points, with mutually
exclusive robust factors whose different means would fight it out
via GBP, in collaboration with other inter-measurement factors
representing smoothness, etc., until the most probably associations
are reached. This could be something which could be efficiently
implemented on a distributed close-to-the-sensor processor.

\section{GBP Convergence Properties and Message Schedules}

In this paper, we have focused on the local details of GBP, and shown
various examples, and it is not within our current scope to examine
the large scale properties of the algorithm. In fact, previously
published implementations and some of our own as-yet unpublished experiments are very
promising, though there are still many open questions.

The most common impression of GBP in extended, loopy linear problems is that
it converges to correct means with overconfident covariances, and this
is certainly what we have observed in our example implementations. 
The covariance overconfidence we observe can be most easily understood in Malioutov \etal's  Walk Sum Framework \cite{Malioutov:etal:2006} in which the loopy graph is unrolled into a computation graph that is a tree.
It is remarkable that the means in loopy GBP converge
to correct values even when the covariances are overconfident \cite{Weiss:Freeman:NIPS2000}. This
means that the precisions on the incoming messages that each variable
receives must be in the correct proportions. Per variable
  node, there must be constant factor by which the incoming
  messages at convergence are stronger than they should be. Can we
  back this observation out to the factors and the whole graph, and
  then understand how to correct for overconfidence?
A clear empirical pattern is that overconfidence is greater for nodes
in densely connected parts of a graph, where messages will travel over
very loopy, self-reinforcing paths. Perhaps some simple heuristics
would be able to describe the behaviour well even without a full theory.
We recall iSAM2 \cite{Kaess:etal:IJRR2012}, where message passing is restricted to the spanning of a graph tree so that marginal covariance estimates are conservative. Maintaining the spanning tree is however not a local operation.

Importantly, it is known that loopy linear GBP does not always converge.
More specifically, previous authors have found that while the marginal
covariances will always converge to the same values given a positive semi-definite message information matrix initialisation \cite{Du:etal:2017}, whether or not
the means will converge depends on the spectral radius of the linear system for the means that is formed after convergence of the covariances. 
It has been shown that message damping can improve convergence in
these situations \cite{Malioutov:etal:2006}. In message damping, at each message passing step the
calculated message is combined in a fixed weighted ratio with the previous
message sent along that particular edge.

Different issues arise in non-linear problems, where factors must be
repeatedly relinearlised. Clearly, one strategy is to delay
relinearisation, such that with a chosen linearisation for each factor
we iterate message passing until convergence, with the same properties
as any linear problem, and then relinearise everything and
repeat. More aggressive local relinearisation is attractive, because
each factor is always using its current estimates to linearise,  but we
are not yet sure of the effect of this on global
convergence. Dynamically changing the factors as we do in our
technique for using robust kernels does not seem to cause any
unstability in optimisation.

When it comes to message schedules for GBP, beyond the first 1D
example in Section~\ref{section:1d} where we showed a sequential
floodfill schedule, we have focused on schedules which can be
implemented in purely parallel architectures, in the simplest way such
that every variable sends messages, then every factor sends messages.

Such a schedule is sometimes obviously inefficient, for instance in
transmitting absolute pose information which is available only at one
node throughout the whole graph, which happens in very slow steps, and
most of the graph is churning away in parallel for many iterations
without much changing.
A good question is to what extent this could be improved on without
needing global knowledge of the structure of the graph. Certainly, at
a local level, individual nodes could decide only to come alive and
pass messages when their incoming messages have changed by a
significant enough amount, perhaps judged using information theoretic
measures~\cite{Davison:ICCV2005}. This is reminiscent of the
`Wildfire' algorithm in Loopy SAM~\cite{Ranganathan:etal:IJCAI2007}.
Information theoretic measures will also be crucial in the longer term
in deciding on the number of bits needs to specify the quantities used
in message passing.

However, thoughts about message schedules should take
into account our interest in GBP as part of Spatial AI, where the
graphs of interest are always dynamically changing; and where factors and
variables may
have many heterogeneous forms and connection patterns. We foresee
systems where messages passing takes place continuously as the graph
grows, changes, and sometimes simplifies, and perhaps never or rarely
reaches global convergence. Factors may exist which have very
long-reaching effects, due to recognition of large structures for
instance, and these may have very good properties for global
estimation even with few iterations. Most fundamentally, we believe
that the `master representation' of our approach should be the factor
graph itself, and that all marginal estimates due to message passing
are ultimately ephemeral, reversible and that most estimates of
interest are recomputable locally if needed. Therefore a standard analysis
of convergence may not be the most important property.

\section{Towards Fully Graph-Based Spatial AI}

Let us get back to the vision we suggested in the
introduction of this paper, where all or most of the storage and processing
for real-time Spatial AI is carried out in a purely distributed way,
suitable for novel architectures such as graph processors, or even
loose constellations of independent communicating
devices. Probabilistic, geometric estimation as enabled by GBP must be
combined with learned recognition capabilities to produce semantically
meaningful representations. There is certainly
great potential for full integration with the increasingly
important and developed area of Geometric Deep Learning~\cite{Bronstein:etal:SPM2017}, which
operates on graph data by design, and the promise of fully graph-based
semantic SLAM systems. Clusters of related, linked nodes can be
recognised as objects or other high level entities by a graph neural
network. New factors which impose the regularity (e.g. smoothness) of
these entities can be added and incorporated into GBP.

If confidence is high enough, recognised node clusters can be simplified into more efficient parametrised
representations, by combining many nodes into object supernodes, or
even hierarchies of these.
We believe that the 
destiny of a Spatial AI graph representation is to become a sparse object
graph as in SLAM++~\cite{Salas-Moreno:etal:CVPR2013}.
In fact, this simplification is essential, for if we are to retain a factor graph as
our master representation of a scene, and store and update it in
real-time on a
practical embedded processor, the size of a graph must remain
bounded. As new measurements from live sensors in a dynamic scene flow
in, the graph must be repeatedly abstracted and simplified to remain
finite.
Recognition of higher level entities could come from
outside the graph (such as from a CNN running on input images). Or,
for processing to
remain truly distributed, it would occur locally, within the graph,
from learned and geometric graph-based measures.

Meanwhile, can local changes and abstraction
lead towards a graph with the right global properties to stay
useful? Ideally, we believe that a Spatial AI graph should have `small
world' properties, such that a path exists between any two nodes which
has a bounded number of edge hops, and therefore that information flow
and optimisation can always happen efficiently. To what extent are we
able to build such a structure into a graph, and for that structure
not to be purely a representation of the simplest compilation of
measurement data?

%- Think like a vertex, think like a graph, etc.

We will consider these issues in a little more detail in the following sections.

\subsection{Recognition, Semantics and Objects}

In GBP, it is trivial to take a set of variable nodes and merge them
into a single supernode. The supernode has a state vector which is the
concatenation of the individual variable states. All factors which
were connected to the individual variables now connect to the
supernode, with appropriate sparse measurement functions and Jacobians
to reference the right variables. Any factors between variables which
have all been joined into the supernode become new unary factors.

However, we do not gain much advantage by just joining nodes like
this; message passing through the supernode becomes expensive because
large matrices must now be manipulated due to its large state and
large number of connections to factors.  The benefits occur if we can
simplify the state representation of the supernode to something much
smaller. For example, imagine a set of variables representing points
on a surface which are joined together because a recognition module
has identitied the surface as a plane. If we are confident enough
about this, we can replace the many point coordinates by the
parametric equation of an infinite plane, which can be represented
with only three parameters (this is reminiscent of~\cite{Salas-Moreno:etal:ISMAR2014}). All factors connected to the supernode
could then be transformed to relate to these parameters.  As it
stands, the supernode would still have a very large number of factors
connecting it to all the other parts of the graph which connected to
the original variables. However, if these other graph parts are
gradually simplified too, factors can be combined together and the
number of connections will also reduce greatly. For instance, if the
planar surface region has been observed by a moving camera over a long
period of time, and each historic camera position is represented by an
individual variable node, in our factor graph each of these camera
variables will be connected to the plane supernode. However, it may be
that the relative locations of the camera variable nodes at some point
become so well known and this section of the graph so locally `rigid'
that we can also replace all of those nodes with a simpler entity such
as a `trajectory segment' supernode, the many factors linking to it to
the plane could be summarised and combined into efficient `superfactors'.

The precise implementation details of such operations are still to
be worked out, but we find this direction very convincing.
Some interesting recent papers on non-linear factor
recovery~\cite{Mazuran:etal:IJRR2016,Usenko:etal:ARXIV2019} may
provide useful tools for summarising parts of a graph with new
superfactors which retain a non-linear form, and this is consistent
with our picture of keeping the factor graph as the master representation.

We should be clear that 
combining nodes into simpler entities is difficult or impossible to reverse,
unlike most of the processing we do in GBP.  However, making greedy
assumptions and simplifying representations must ultimately be an
essential part of efficient intelligence, and something that
biological brains do as proven by optical illusions.

\subsection{Overall Structure}

A related area for research, as graphs grow larger with more
measurements and variables, but also incrementally become abstracted
and simplified, is how the whole graph should be laid out with respect
to computing hardware, and whether the whole graph structure enables
efficient global or just-in-time inference of properties of interest.

On a graph processor such as Graphcore's IPU, which has distributed
graph computation but is still at heart a synchronous processor with
global clocking and coordination,
a significant issue
is that the structure of the computation graph must  normally be decided and
fixed at compile time. With dynamic Spatial AI graphs, the most
obvious option is to precompile a graph structure which we estimate is `big
enough' for the problem of interest, and then to fill that structure
up dynamically at run-time. We believe that this is both feasible and
practical. We could perform analysis of the typical graphs we would
ideally build in an off-line experiment, and measure the typical
number of nodes and amount of interconnectivity and use these to
choose parameters of the pre-defined graph.
Of course this need for graph compilation may go away with alternative future hardware.

Another interesting issue relates to our observation (e.g. in Figure~\ref{fig:2dtest}) that in SLAM-like problems with mainly relative measurements, GBP will often produce good relative estimates rather quickly while it can take many iterations are needed to get good absolute estimates. In many applications, it is only the relative information which is important (for a robot which needs to plan its next actions to avoid obstacles for instance), and this raises the question of whether a different parametrisation which is relative by definition would be more suitable. This question has been considered before, such as in Sibley \etal's Relative Bundle Adjustment \cite{Sibley:etal:RSS2009}, where all scene points were represented relative to a camera pose, and loop closure could happen as a simple connection operation with the option for global adjustment only if needed. These methods deserve being looked at again, though we wonder still how well they generalise to general graphs and motion rather than the `corridor-like' trajectories of cameras they were originally devised for.

\subsection{Active Processing and Attention}

If a whole graph is stored within a graph processor, it can be
operated on via GBP `in place', and potentially all parts of the graph
could circulate messages at the same rate (as in our SLAM examples). However, it is unlikely
that this would make sense in terms of power efficiency, and that
there are likely to be parts of the graph which are currently of much
more interest which will be a priority for processing --- most
obviously, at the current location of a moving device or robot which
is making current sensor measurements and must decide on
action. Processing `attention' could actively be focused on this area
with a high rate of message passing, while other parts of the graph
are partially or completely neglected, to be picked up and updated
later on as needed `just in time'. We imagine an attention spotlight
which moves around the graph, bringing it to life.

Depending on memory constraints (and current graph processors do not
have a huge amount of on chip memory, e.g. around 300MB for one
Graphcore IPU), graph regions out of the current attention spotlight might
even be much abstracted and simplified to low resolution, approximated
forms maintaining only the main shape and connectivity, perhaps in an
analogue of the
way that a human brain remembers distant places. When a moving device
with data-rich sensors such as cameras revisits these places, they can
easily feed on this data to be brought back to the high resolution
needed for local action.

There may need to be a particular region of a graph processor held
aside as the current active workspace, where enough precompiled space
is retained such that the live part of the graph can be copied, unpacked and
subject to full rate processing
(the Real-Time Loop part of the `Spatial AI Brain' shown in Figure 4 of~\cite{Davison:ARXIV2018}).
Major processing elements such as a semantic labelling deep network
could be held to run permanently here, only to operate on live image
data (if such a network has many weights then it would not be possible
in any case to distribute many copies of those weights around the
whole graph so that labelling could happen in any location).

Between the active workspace and the rest of the graph there will need
to be some special graph infrastructure such as routing nodes to
interface between live workspace and long-term graph memory.

\subsection{Graphs of Multiple Communicating Devices}

In the final part of this speculative section, we consider the
strong prospects for GBP methods to be used outside of the confines of a
single graph processor, to connect many independent but
intercommunicating devices such that they can jointly estimate global
quantities. 
Individual devices will have their own individual estimation
algorithms, sensors and hardware, but could use GBP as the general
`glue language' to share probabilistic information and come to
agreement over global estimation matters, such as if a swarm of
robotic devices were to be organised into a regular grid via only
local computation and communication.

What will be important to achieve such capabilities will be standards
for interoperation and messages which can be deployed for
communication.
Inspired by the creation of the World Wide Web
\cite{BernersLee:Book1999}, we believe that a fairly limited set of open
communication standards which are the equivalent of HTTP, HTML, URL,
etc. could define how devices could send probabilistic messages to
each other while each maintains and runs its own internal operation in
whichever proprietory way is desired.
Such a set of standards could be the most important way that GBP
becomes influential, as a standard for distributed estimation, rather than it being instantiated in monolithic
libraries like other estimation methods.

%% In his book `Out of Control: The New Biology of Machines, Social
%% Systems and the Economic World' 
%% Big AI systems will be more like biological entities
%% \cite{Kelly:Book1992}

%% In natural systems, improvements are “pasted” over an existing
%% debugged system. The original layer continues to operate without even
%% being (or needing to be) aware that is has another layer above
%% it. (Subsumption \cite{Brooks:WFAI1987})

\section{Conclusions}

We have presented a detailed case for a serious reconsideration of
Gaussian Belief Propagation in Spatial AI, inspired chiefly by the
current rapid developments in processor hardware which open up the
chance to close the large gap between the requirements for advanced
perception on embodied intelligent products and what can be delivered
with practical mass-market, low-power technology. The advantage will
increasingly be with algorithms which can
take advantage of purely distributed processing and storage but still
deliver globally meaningful estimates of the properties of a device
and its surroundings. 

A non-linear factor graph is the purest representation of probabilistic
knowledge from multiple heterogeneous information sources, and can be
efficiently stored and dynamically edited as the master representation
for Spatial AI. Gaussian Belief Propagation is the simple but highly
flexible tool which can turn a factor graph into a set of marginal
probabilistic estimates with flexible distributed processing and
storage, suitable either for graph processor chips or networks of
individual devices with pairwise communication.
We imagine large real-time systems operating with continual or attention-focused
processing on their dynamic factor graphs, perhaps never reaching full
estimation convergence but with estimates always good enough to be
useful, either locally or globally, or intensively calculated on demand
in a just-in-time manner. More specifically designed algorithms and
specialised computing hardware for these will always exist for
focused uses, but GBP can serve as a general `glue' which holds all
of these together in a rigorous probabilistic framework. An important
systems focus will be the definition of interfaces which allow
multiple devices to pass messages between them.

The factor graph in such a system must be kept bounded, and we believe
that the route towards this is continual graph-based introspection to
discover regions and structures which can be simplified by deleting or
merging nodes. These mechanisms should interface with learned
graph-based recognition mechanisms which identify and segment objects
or simple structures which can be efficiently parameterised.

There are clearly many open questions in this research area,
especially related to the convergence of estimation in large graphs,
where there is much research and new theories needed. We have enough confidence in the potential of GBP methods for Spatial
AI to believe that it is worth spending the many years needed on the research to
take them into practical systems. This is an algorithmic framework ready
not just for processor types already in production, but even more
exotic future possibilities such as neuromorphic devices which give up
on global timing and synchronisation.

\section*{Acknowledgements}

We are grateful to many researchers with whom we have discussed some
of the ideas in this paper, especially from the Dyson Robotics Lab and
Robot Vision Group at Imperial College, and SLAMcore. We would
particularly like to thank Stefan
Leutenegger, Jan Czarnowski, Paul Kelly, Jacek Zienkiewicz, Owen
Nicholson, Julien Martel, Pablo Alcantarilla, Juan Tarrio, Ujwal Bonde, Maxime Boucher, Alexandre Morgand,
Richard Newcombe,  Michael Bloesch, Ronnie Clark, Sajad Saeedi, Simon Knowles, Mark Pupilli, Tristan
Laidlow, Edgar
Sucar, Kentaro Wada, Shuaifeng Zhi, Charles Collis, Rob Deaves, Margarita Chli, Panos Parpas, Mahdi Cheraghchi,
Piotr Dudek, Walterio Mayol-Cuevas, Steve Furber,
Robert Mahony, Frank Dellaert and Ian Reid.

%%%%%%%%% REFERENCES
{\small
\bibliographystyle{ieee}
\bibliography{robotvision}

\begin{thebibliography}{10}\itemsep=-1pt

\bibitem{Graphcore}
Graphcore.
\newblock URL https://www.graphcore.ai/.

\bibitem{Agarwal:etal:ICRA2013}
P.~Agarwal, G.~D. Tipaldi, L.~Spinello, C.~Stachniss, and W.~Burgard.
\newblock Robust map optimization using dynamic covariance scaling.
\newblock In {\em {Proceedings of the {IEEE} International Conference on
  Robotics and Automation ({ICRA})}}, 2012.

\bibitem{CeresManual}
S.~Agarwal, M.~K., and Others.
\newblock Ceres solver.
\newblock \url{http://ceres-solver.org}.

\bibitem{BernersLee:Book1999}
T.~Berners-Lee.
\newblock {\em {Weaving the Web: The Original Design and Ultimate Destiny of
  the World Wide Web}}.
\newblock Harper, 1999.

\bibitem{Bishop:Book2006}
C.~M. Bishop.
\newblock {\em {Pattern Recognition and Machine Learning}}.
\newblock Springer-Verlag New York, Inc., 2006.

\bibitem{Bloesch:etal:CVPR2018}
M.~Bloesch, J.~Czarnowski, R.~Clark, S.~Leutenegger, and A.~J. Davison.
\newblock {CodeSLAM} --- learning a compact, optimisable representation for
  dense visual {SLAM}.
\newblock In {\em {Proceedings of the {IEEE} Conference on Computer Vision and
  Pattern Recognition ({CVPR})}}, 2018.

\bibitem{Bronstein:etal:SPM2017}
M.~M. Bronstein, J.~Bruna, Y.~LeCun, A.~Szlam, and P.~Vandergheynst.
\newblock Geometric {D}eep {L}earning: {G}oing beyond {E}uclidean data.
\newblock {\em IEEE Signal Processing Magazine}, 34(4):18--42, 2017.

\bibitem{Chambolle:2011}
A.~Chambolle and T.~Pock.
\newblock {A First-Order Primal-Dual Algorithm for Convex Problems with
  Applications to Imaging}.
\newblock {\em {Journal of Mathematical Imaging and Vision}}, 40(1):120--145,
  2011.

\bibitem{Choudhary:etal:IJRR2017}
S.~Choudhary, L.~Carlone, C.~Nieto, J.~Rogers, H.~Christensen, and F.~Dellaert.
\newblock Distributed mapping with privacy and communication constraints:
  Lightweight algorithms and object-based models.
\newblock {\em {International Journal of Robotics Research ({IJRR})}},
  36(12):1286--1311, 2017.

\bibitem{Crandall:etal:CVPR2011}
D.~Crandall, A.~Owens, N.~Snavely, and D.~Huttenlocher.
\newblock Discrete-continuous optimization for large-scale structure from
  motion.
\newblock In {\em {Proceedings of the {IEEE} Conference on Computer Vision and
  Pattern Recognition ({CVPR})}}, 2011.

\bibitem{Davison:ICCV2005}
A.~J. Davison.
\newblock {Active Search for Real-Time Vision}.
\newblock In {\em {Proceedings of the International Conference on Computer
  Vision ({ICCV})}}, 2005.

\bibitem{Davison:ARXIV2018}
A.~J. Davison.
\newblock {FutureMapping}: The computational structure of {Spatial AI} systems.
\newblock {\em arXiv preprint arXiv:arXiv:1803.11288}, 2018.

\bibitem{Dellaert:Kaess:Foundations2017}
F.~Dellaert and M.~Kaess.
\newblock {Factor Graphs for Robot Perception}.
\newblock {\em Foundations and Trends in Robotics}, 6(1--2):1--139, 2017.

\bibitem{Du:etal:2017}
J.~Du, S.~Ma, Y.-C. Wu, S.~Kar, and J.~M. Moura.
\newblock Convergence analysis of distributed inference with vector-valued
  gaussian belief propagation.
\newblock {\em Journal of Machine Learning Research}, 18:172--1, 2017.

\bibitem{Engel:PhDThesis:2017}
J.-J. Engel.
\newblock {\em Large-Scale Direct SLAM and 3D Reconstruction in Real-Time}.
\newblock PhD thesis, 2017.

\bibitem{Eustice:etal:ICRA2005}
R.~M. Eustice, H.~Singh, and J.~J. Leonard.
\newblock {Exactly Sparse Delayed State Filters}.
\newblock In {\em {Proceedings of the {IEEE} International Conference on
  Robotics and Automation ({ICRA})}}, 2005.

\bibitem{Grisetti:etal:ITITS2009}
G.~Grisetti, C.~Stachniss, and W.~Burgard.
\newblock Non-linear constraint network optimization for efficient map
  learning.
\newblock {\em {IEEE} Transactions on Intelligent Transportation Systems},
  10(3):428--439, 2009.

\bibitem{Jaynes:Book2003}
E.~T. Jaynes.
\newblock {\em {Probability Theory: The Logic of Science}}.
\newblock Cambridge University Press, 2003.

\bibitem{Kaess:etal:IJRR2012}
M.~Kaess, H.~Johannsson, R.~Roberts, V.~Ila, J.~Leonard, and F.~Dellaert.
\newblock {{iSAM2}: Incremental Smoothing and Mapping Using the {B}ayes Tree}.
\newblock {\em {International Journal of Robotics Research ({IJRR})}}, 2012.
\newblock To appear.

\bibitem{Kummerle:etal:ICRA2011}
R.~K{\"u}mmerle, G.~Grisetti, H.~Strasdat, K.~Konolige, and W.~Burgard.
\newblock {{$g^2o$}: A General Framework for Graph Optimization}.
\newblock In {\em {Proceedings of the {IEEE} International Conference on
  Robotics and Automation ({ICRA})}}, 2011.

\bibitem{Leutenegger:etal:IJRR2014}
S.~Leutenegger, S.~Lynen, M.~Bosse, R.~Siegwart, and P.~Furgale.
\newblock Keyframe-based visual-inertial odometry using nonlinear optimization.
\newblock {\em The International Journal of Robotics Research}, 34(3):314--334,
  2014.

\bibitem{Malioutov:etal:2006}
D.~M. Malioutov, J.~K. Johnson, and A.~S. Willsky.
\newblock {Walk-sums and belief propagation in Gaussian graphical models}.
\newblock {\em Journal of Machine Learning Research}, 7(Oct):2031--2064, 2006.

\bibitem{Martel:PHD2019}
J.~Martel.
\newblock {\em Unconventional Processing with Unconventional Visual Sensing}.
\newblock PhD thesis, {ETH Zurich}, 2019.

\bibitem{Mazuran:etal:IJRR2016}
M.~Mazuran, W.~Burgard, and G.~D. Tipaldi.
\newblock Nonlinear factor recovery for long-term {SLAM}.
\newblock {\em {International Journal of Robotics Research ({IJRR})}},
  35(1):50--72, 2016.

\bibitem{McCormac:etal:ICRA2017}
J.~McCormac, A.~Handa, A.~J. Davison, and S.~Leutenegger.
\newblock {SemanticFusion}: Dense {3D} semantic mapping with convolutional
  neural networks.
\newblock In {\em {Proceedings of the {IEEE} International Conference on
  Robotics and Automation ({ICRA})}}, 2017.

\bibitem{Mourikis:Roumeliotis:ICRA2007}
A.~I. Mourikis and S.~I. Roumeliotis.
\newblock A multi-state constraint {Kalman} filter for vision-aided inertial
  navigation.
\newblock In {\em Robotics and Automation, 2007 IEEE International Conference
  on}, pages 3565--3572. IEEE, 2007.

\bibitem{Newcombe:PHD2012}
R.~A. Newcombe.
\newblock {\em {Dense Visual {SLAM}}}.
\newblock PhD thesis, Imperial College London, 2012.

\bibitem{Newcombe:etal:ISMAR2011}
R.~A. Newcombe, S.~Izadi, O.~Hilliges, D.~Molyneaux, D.~Kim, A.~J. Davison,
  P.~Kohli, J.~Shotton, S.~Hodges, and A.~Fitzgibbon.
\newblock {{KinectFusion}: Real-Time Dense Surface Mapping and Tracking}.
\newblock In {\em {Proceedings of the International Symposium on Mixed and
  Augmented Reality ({ISMAR})}}, 2011.

\bibitem{Olson:Agarwal:IJRR2013}
E.~Olson and P.~Agarwal.
\newblock Inference on networks of mixtures for robust robot mapping.
\newblock {\em {International Journal of Robotics Research ({IJRR})}},
  32(7):826--840, 2013.

\bibitem{Olson:etal:ICRA2006}
E.~Olson, J.~J. Leonard, and S.~Teller.
\newblock {Fast iterative alignment of pose graphs with poor initial
  estimates}.
\newblock In {\em {Proceedings of the {IEEE} International Conference on
  Robotics and Automation ({ICRA})}}, 2006.

\bibitem{Paskin:IJCAI2003}
M.~A. Paskin.
\newblock {Thin Junction Tree Filters for Simultaneous Localization and
  Mapping}.
\newblock In {\em {Proceedings of the International Joint Conference on
  Artificial Intelligence ({IJCAI})}}, 2003.

\bibitem{Pearl:AAAI1982}
J.~Pearl.
\newblock {Reverend Bayes} on inference engines: A distributed hierarchical
  approach.
\newblock In {\em {Proceedings of the National Conference on Artificial
  Intelligence ({AAAI})}}, 1982.

\bibitem{Ranganathan:etal:IJCAI2007}
A.~Ranganathan, M.~Kaess, and F.~Dellaert.
\newblock {Loopy SAM}.
\newblock In {\em {Proceedings of the International Joint Conference on
  Artificial Intelligence ({IJCAI})}}, 2007.

\bibitem{Saeedi:etal:JFR2016}
S.~Saeedi, M.~Trentini, M.~Seto, and H.~Li.
\newblock {Multiple-robot Simultaneous Localization and Mapping --- A Review}.
\newblock {\em {Journal of Field Robotics ({JFR})}}, 33(1):3--46, 2016.

\bibitem{Salas-Moreno:etal:ISMAR2014}
R.~F. Salas-Moreno, B.~Glocker, P.~H.~J. Kelly, and A.~J. Davison.
\newblock Dense planar {SLAM}.
\newblock In {\em {Proceedings of the International Symposium on Mixed and
  Augmented Reality ({ISMAR})}}, 2014.

\bibitem{Salas-Moreno:etal:CVPR2013}
R.~F. Salas-Moreno, R.~A. Newcombe, H.~Strasdat, P.~H.~J. Kelly, and A.~J.
  Davison.
\newblock {{SLAM++}: Simultaneous Localisation and Mapping at the Level of
  Objects}.
\newblock In {\em {Proceedings of the {IEEE} Conference on Computer Vision and
  Pattern Recognition ({CVPR})}}, 2013.

\bibitem{Sibley:etal:RSS2009}
G.~Sibley, C.~Mei, I.~Reid, and P.~Newman.
\newblock {Adaptive Relative Bundle Adjustment}.
\newblock In {\em {Proceedings of Robotics: Science and Systems ({RSS})}},
  2009.

\bibitem{Strasdat:etal:ICRA2010}
H.~Strasdat, J.~M.~M. Montiel, and A.~J. Davison.
\newblock {Real-Time Monocular {SLAM}: Why Filter?}
\newblock In {\em {Proceedings of the {IEEE} International Conference on
  Robotics and Automation ({ICRA})}}, 2010.

\bibitem{Sudderth:etal:TSP2004}
E.~B. Sudderth, M.~J. Wainwright, and A.~S. Willsky.
\newblock {Embedded trees: Estimation of Gaussian processes on graphs with
  cycles}.
\newblock {\em {IEEE Transactions on Signal Processing}}, 52(11):3136--3150,
  2004.

\bibitem{Sutter:Jungle2011}
H.~Sutter.
\newblock Welcome to the jungle.
\newblock URL https://herbsutter.com/welcome-to-the-jungle, 2011.

\bibitem{Triggs:etal:VISALG1999}
B.~Triggs, P.~McLauchlan, R.~Hartley, and A.~Fitzgibbon.
\newblock {Bundle Adjustment --- A Modern Synthesis}.
\newblock In {\em {Proceedings of the International Workshop on Vision
  Algorithms, in association with {ICCV}}}, 1999.

\bibitem{Usenko:etal:ARXIV2019}
V.~Usenko, N.~Demmel, D.~Schubert, J.~St\"{u}ckler, and D.~Cremers.
\newblock Visual-inertial mapping with non-linear factor recovery.
\newblock {\em arXiv preprint:1904:11781}, 2019.

\bibitem{Weerasekera:etal:ICRA2018}
C.~S. Weerasekera, T.~Dharmasiri, R.~Garg, T.~Drummond, and I.~D. Reid.
\newblock {Just-in-Time Reconstruction:} inpainting sparse maps using single
  view depth predictors as priors.
\newblock In {\em {Proceedings of the {IEEE} International Conference on
  Robotics and Automation ({ICRA})}}, 2018.

\bibitem{Weiss:Freeman:NIPS2000}
Y.~Weiss and W.~T. Freeman.
\newblock {C}orrectness of belief propagation in {G}aussian graphical models of
  arbitrary topology.
\newblock In {\em {Neural Information Processing Systems ({NIPS})}}, pages
  673--679, 2000.

\bibitem{Whelan:etal:RSS2015}
T.~Whelan, S.~Leutenegger, R.~F. Salas-Moreno, B.~Glocker, and A.~J. Davison.
\newblock {ElasticFusion}: Dense {SLAM} without a pose graph.
\newblock In {\em {Proceedings of Robotics: Science and Systems ({RSS})}},
  2015.

\bibitem{Wu:etal:BA:CVPR2011}
C.~Wu, S.~Agarwal, B.~Curless, and S.~M. Seitz.
\newblock {Multicore Bundle Adjustment}.
\newblock In {\em {Proceedings of the {IEEE} Conference on Computer Vision and
  Pattern Recognition ({CVPR})}}, 2011.

\bibitem{Zhi:etal:CVPR2019}
S.~Zhi, M.~Bloesch, S.~Leutenegger, and A.~J. Davison.
\newblock {SceneCode}: Monocular dense semantic reconstruction using learned
  encoded scene representations.
\newblock In {\em {Proceedings of the {IEEE} Conference on Computer Vision and
  Pattern Recognition ({CVPR})}}, 2019.

\bibitem{Zienkiewicz:PHD2017}
J.~Zienkiewicz.
\newblock {\em {Dense Monocular Perception for Mobile Robotics}}.
\newblock PhD thesis, Imperial College London, 2017.

\end{thebibliography}
}

\end{document}